\pdfoutput=1

\documentclass[11pt]{article}
\PassOptionsToPackage{dvipsnames}{xcolor}

\usepackage[]{acl}

\usepackage{times}
\usepackage{latexsym}

\usepackage[T1]{fontenc}

\usepackage[utf8]{inputenc}

\usepackage{microtype}

\usepackage{algorithm, algorithmic, letltxmacro}
\usepackage{amsmath}
\usepackage{amssymb}
\usepackage{amsfonts}
\usepackage{pgfplots}
\usetikzlibrary[angles, arrows.meta, decorations.pathmorphing, decorations.pathreplacing, backgrounds, intersections, positioning, fit, petri, graphs, math, external, calc, shapes, shapes.geometric, decorations.text, shadings, pgfplots.groupplots]
\usepackage{graphicx}
\usepackage{tikz}
\usepackage{tikz-dependency}
\usepackage{subcaption}
\usepackage{multirow}
\usepackage{booktabs}
\usepackage{mathtools}
\usepackage[dvipsnames]{xcolor}
\usepackage{enumitem}
\usepackage{mathpartir}
\usepackage{mathabx}
\usepackage{tablefootnote}
\usepackage{colortbl}
\usepackage{arydshln}
\usepackage{CJKutf8}
\usepackage{forest}
\usepackage{latexsym}
\usepackage{mathabx}
\usepackage{stmaryrd}
\usepackage{wasysym}
\usepackage{fontawesome}
\forestset{
dg edges/.style={for tree={parent anchor=south, child anchor=north,align=center,base=bottom,where n children=0{tier=word,edge=dotted,calign with current edge}{}}},
}

\setlist{nosep}

\definecolor{brickred}{HTML}{b92622}
\definecolor{midnightblue}{HTML}{005c7f}
\definecolor{salmon}{HTML}{f1958d}
\definecolor{burntorange}{HTML}{f19249}
\definecolor{junglegreen}{HTML}{4dae9d}
\definecolor{forestgreen}{HTML}{499c5e}
\definecolor{pinegreen}{HTML}{3d8a75}
\definecolor{seagreen}{HTML}{6bc1a2}
\definecolor{limegreen}{HTML}{97c65a}
\newcommand{\white}[1]{\textcolor{white}{#1}}
\newcommand{\brickred}[1]{\textcolor{brickred}{#1}}

\newcommand*{\rulefiller}{%
  \arrayrulecolor[gray]{0.95}%
  \specialrule{\heavyrulewidth}{0pt}{-\heavyrulewidth}%
  \arrayrulecolor{black}%
}
\newcommand{\suda}{\textsuperscript{\faStarO}}
\newcommand{\huawei}{\textsuperscript{\faMoonO}}
\newcommand{\damo}{\textsuperscript{\faSunO}}

\DeclareDocumentCommand{\trapezoid}{O{2.0} O{1.0} O{0.5} m m m}{
  \begin{scope}[scale=0.9,thick]
    \draw[anchor=mid] (0, 0) -- (0, -{#2}) node[below=7.5pt,anchor=base] {\footnotesize\ensuremath{#4}} -- ({#1}, -{#2}) node [below=7.5pt,anchor=base] {\ensuremath{#5}} -- ({#1}, -{#3}) -- cycle;
    \draw node[] at ($(0, 0)!0.5!({#1}, {#3}) + (0, 0.2)$) {\footnotesize\ensuremath{#6}};
  \end{scope}
}
\DeclareDocumentCommand{\square}{O{0.5} O{0.5} m m m}{
  \begin{scope}[scale=0.9,thick]
    \draw[anchor=mid] (0, -{#2}) node[below left=7.5pt and 3pt,anchor=base] {\footnotesize\ensuremath{#3}} -- ({#1}, -{#2}) node (square) [below=7.5pt,anchor=base] {\footnotesize\ensuremath{#4}} -- ({#1}, 0) --  (0, 0) -- cycle;
    \draw node[] at ($(0, 0)!0.5!({#1}, 0) + (0, 0.2)$) {\footnotesize\ensuremath{#5}};
  \end{scope}
}
\DeclareDocumentCommand{\lefttriangle}{O{0.5} O{0.5} m m m}{
  \begin{scope}[scale=0.9,thick]
    \draw[anchor=mid] (0, -{#2}) node[below left=7.5pt and 3pt,anchor=base] {\footnotesize\ensuremath{#3}} -- ({#1}, -{#2}) node [below=7.5pt,anchor=base] {\footnotesize\ensuremath{#4}} -- ({#1}, 0) -- cycle;
    \draw node[] at ($(0, -{#2})!0.5!({#1}, 0) + (0, 0.45)$) {\footnotesize\ensuremath{#5}};
  \end{scope}
}
\DeclareDocumentCommand{\righttriangle}{O{0.5} O{0.5} m m m}{
  \begin{scope}[scale=0.9,thick]
    \draw[anchor=mid](0, 0) -- (0, -{#2}) node[below=7.5pt,anchor=base] {\footnotesize\ensuremath{#3}} -- ({#1}, -{#2}) node [below right=7.5pt and 3pt,anchor=base] {\footnotesize\ensuremath{#4}} -- cycle;
    \draw node[] at ($(0, 0)!0.5!({#1}, -{#2}) + (0, 0.45)$) {\footnotesize\ensuremath{#5}};
  \end{scope}
}

\title{Semantic Role Labeling as Dependency Parsing: Exploring \\
Latent Tree Structures Inside Arguments}

\author{
    \textbf{Yu Zhang}\suda,
    \textbf{Qingrong Xia}\suda\huawei,
    \textbf{Shilin Zhou}\suda,
    \textbf{Yong Jiang}\damo,
    \textbf{Guohong Fu}\suda\thanks{$~~$Corresponding author},
    \textbf{Min Zhang}\suda \\
    \suda Institute of Artificial Intelligence, School of Computer Science and Technology, \\
    Soochow University, Suzhou, China\\
    \huawei Huawei Cloud, China \\
    \damo DAMO Academy, Alibaba Group, China \\
    \texttt{\{yzhang.cs,slzhou.cs\}@outlook.com; xiaqingrong@huawei.com} \\
    \texttt{yongjiang.jy@alibaba-inc.com; \{ghfu,minzhang\}@suda.edu.cn}
}

\date{\today}

\begin{document}
\maketitle
\begin{abstract}
    Semantic role labeling (SRL) is a fundamental yet challenging task in the NLP community.
    Recent works of SRL mainly fall into two lines: 1) BIO-based; 2) span-based.
    Despite ubiquity, they share some intrinsic drawbacks of not considering internal argument structures, potentially hindering the model's expressiveness.
    The key challenge is arguments are flat structures, and there are no determined subtree realizations for words inside arguments.
    To remedy this, in this paper, we propose to regard flat argument spans as latent subtrees, accordingly reducing SRL to a tree parsing task.
    In particular, we equip our formulation with a novel span-constrained TreeCRF to make tree structures span-aware and further extend it to the second-order case.
    We conduct extensive experiments on CoNLL05 and CoNLL12 benchmarks.
    Results reveal that our methods perform favorably better than all previous syntax-agnostic works, achieving new state-of-the-art under both \emph{end-to-end} and \emph{w/ gold predicates} settings.
\end{abstract}

\section{Introduction}\label{sec:intro}

Semantic role labeling (SRL) is a fundamental yet challenging task in the NLP community, involving predicate and argument identification, as well as semantic role classification.
As SRL can provide informative linguistic representations, it has been widely adopted in downstream tasks like question answering \cite{berant-etal-2013-semantic,yih-etal-2016-value}, information extraction \cite{christensen-etal-2010-semantic,lin-etal-2017-neural}, and machine translation \cite{liu-gildea-2010-semantic,bazrafshan-gildea-2013-semantic}, etc.

Recent works of SRL mainly fall into two lines: 1) BIO-based; 2) span-based.
The former views SRL as a sequence labeling task \cite{zhou-xu-2015-end,strubell-etal-2018-lisa,shi-etal-2019-simple}.
For each predicate, each token is tagged with a label starting with BIO prefixes indicating if it is at the \textbf{B}eginning, \textbf{I}nside, or \textbf{O}utside of an argument.
The latter \cite{he-etal-2018-jointly,ouchi-etal-2018-span,li-etal-2019-dependency}, in contrast, opts to jointly predict all predicate and argument span pairs using a span-graph formulation.

\begin{figure}
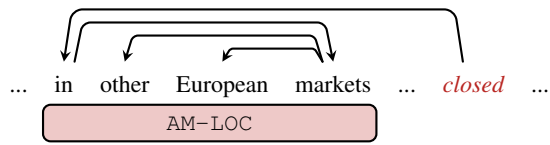

    \centering
    \begin{dependency}
        \begin{deptext}[column sep=0.18cm,font=\small]
            ... \& in \& other \& European \& markets \& ...  \& \emph{\brickred{closed}} \& ... \\
        \end{deptext}
        \depedge[edge vertical padding=0.3ex, hide label, edge height=4ex, thick]{7}{2}{\small NULL}
        \depedge[edge vertical padding=0.3ex, hide label, edge height=2ex, thick]{5}{3}{\small NULL}
        \depedge[edge vertical padding=0.3ex, hide label, edge height=1ex, thick]{5}{4}{\small NULL}
        \depedge[edge vertical padding=0.3ex, hide label, edge height=3ex, thick]{2}{5}{\small NULL}
        \node (a01l) [below left of = \wordref{1}{2}, xshift=2.5ex, yshift=2.5ex] {};
        \node (a01r) [below right of = \wordref{1}{5}, xshift=-0.8ex, yshift=-0.3ex] {};
        \draw [fill=brickred!25, thick, rounded corners=1mm] (a01l) rectangle (a01r);
        \draw [draw=none] (a01l) -- node[] {\small\texttt{AM-LOC}} (a01r);
    \end{dependency}
    \caption{An argument example (below) and its related subtree structure (above) for the predicate ``\brickred{\emph{closed}}''.}
    \label{fig:example}
\end{figure}

Despite ubiquity, there are some drawbacks that limit the expressiveness of the two methods.
First, framing predicate-argument structures as a BIO-tagging scheme is less effective as it lacks explicit modeling of span-level representations, so that long adjacencies of argument phrases can be ignored \cite{cohn-blunsom-2005-semantic,jie-lu-2019-dependency,zhou-etal-2020-latent,xu-etal-2021-better}.
Second, span-based method seeks to pick very few (typically $<$10\%) positive examples from $O(n^3)$ candidate predicate-argument pairs, thus suffering from severe class imbalance problem \cite{li-etal-2021-syntax}.
To alleviate this issue, span-based method relies on heavy pruning \cite{he-etal-2018-jointly} to reduce the searching space, potentially impairing the performance.

Meanwhile, both formulations share some common flaws in terms of lacking explicit modeling of internal argument structures, which appear to be beneficial to SRL.
Taking Fig.~\ref{fig:example} as an example, internal dependencies of words (``in other European markets'') inside the span provide strong clues for recognizing it as a locative modifier (``\texttt{AM-LOC}'') of the predicate ``\emph{\brickred{closed}}''.
Besides, the predicate-argument relation can be naturally reflected by the dependency from the predicate to the span headword (``\emph{\brickred{closed}} $\xrightarrow{\texttt{AM-LOC}}$ in''), and we can properly recognize the argument span boundaries by retrieving all descendants of the subtree.
Such observations have motivated many attempts on utilizing relations inside arguments \cite[\emph{inter alia}]{gildea-hockenmaier-2003-identifying,johansson-nugues-2008-dependency,johansson-nugues-2008-effect,xia-etal-2019-syntax,li-etal-2019-dependency}.
However, stuck on the fact that span-style SRL has no determined internal structure realizations, existing works have to resort to making use of external human-annotated syntax knowledge to bridge the gap \cite{shi-etal-2020-semantic,li-etal-2021-syntax}.

Our main goal in this work is to explicitly take internal argument structures into account meanwhile keeping our framework \emph{end-to-end}.
To this end, we propose to model flat arguments as latent subtrees, thus paving the way for reducing SRL to dependency parsing seamlessly: \emph{we view predicate-argument structures as partially-observed trees where exact subtrees for each argument are not realized yet.}
In this way, we reframe span-style SRL as parsing word-to-word relations by encoding all predicate-argument relations into a unified dependency graph.
Unlike span-based methods \cite{he-etal-2018-jointly}, a dependency graph contains no more than $O(n^2)$ possible dependencies, so that the class imbalance issue can be side-stepped effortlessly.
Specifically, we make use of TreeCRF \cite{eisner-2000-bilexical,zhang-etal-2020-efficient}, which provides a viable way for probabilistic modeling of tree structures, to learn the partially-observed trees and marginalize the latent structures out during training.
Unlike canonical TreeCRF, which enumerates all possible trees, in our setting, we have to impose many span constraints to reflect the argument boundaries on subtrees correctly.
To accommodate this, we further design a novel span-constrained TreeCRF to adapt it to our learning procedure, which explicitly prohibits invalid edges across different arguments as well as multi-head subtrees \cite{nivre-etal-2014-squibs,zhang-etal-2021-adapting}.

There are further advantages to our reduction.
Conversion to tree structures enables us to easily conduct global optimization \cite{eisner-1996-three,mcdonald-etal-2005-online} in polynomial time, which has already been shown to often lead to improved results and more meaningful predictions \cite{toutanova-etal-2008-global, tackstrom-etal-2015-efficient,fitzgerald-etal-2015-semantic,li-etal-2020-structured} compared to local unconstrained methods.
On the other hand, by drawing on the experience in the parsing literature, we can further extend our method to some well-studied high-order methods \cite{mcdonald-pereira-2006-online} without any obstacle.
We experiment with sibling factors in this work and find significant gains, in line with many parsing works \cite{zhang-etal-2020-efficient,fonseca-martins-2020-revisiting}.
Our contributions can be summarized as follows:\footnote{Our code is publicly available at \url{https://github.com/yzhangcs/crfsrl}.}
\begin{itemize}[leftmargin=11pt]
    \item Aware of the benefits of internal argument structures, we propose to model flat argument spans as latent subtrees, thereby reducing SRL to dependency parsing seamlessly.
    \item We propose a novel span-constrained TreeCRF to learn the converted trees and further extend it to the second-order case.
    \item Experiments on CoNLL05 and CoNLL12 benchmarks reveal that our proposed methods outperform existing works significantly, achieving new state-of-the-art results under the syntax-agnostic setting.
\end{itemize}

\section{Overview}\label{sec:srl-as-dep}

\begin{figure*}[th!]
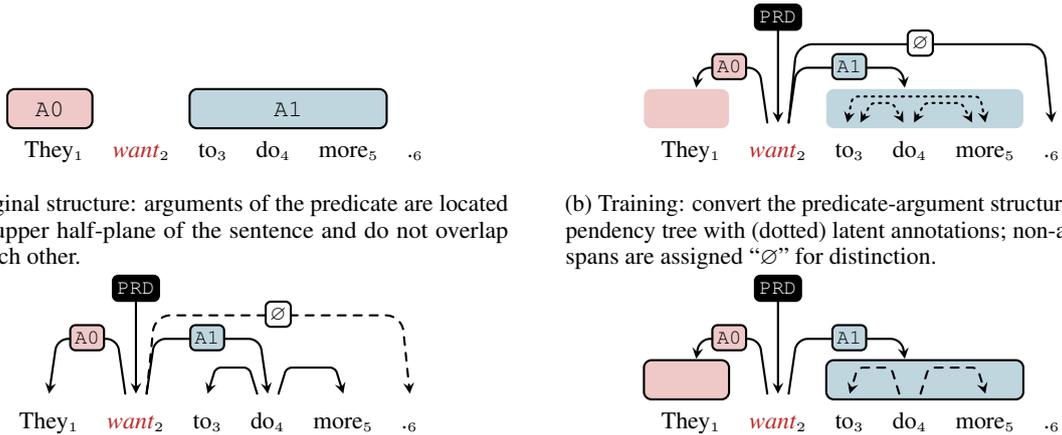

    \begin{subfigure}[b]{0.99\columnwidth}
        \centering
        \begin{dependency}
            \begin{deptext}[column sep=0.2cm,font=\small]
                They\scriptsize$_1$ \& \brickred{\emph{want}}\scriptsize$_2$ \& to\scriptsize$_3$ \& do\scriptsize$_4$ \& more\scriptsize$_5$ \&.\scriptsize$_6$ \\
            \end{deptext}
            \node (a00l) [above left of = \wordref{1}{1}, xshift=0.5ex, yshift=-2.5ex] {};
            \node (a00r) [above right of = \wordref{1}{1}, xshift=-1.2ex, yshift=0.5ex] {};
            \draw [fill=brickred!25, thick, rounded corners=1mm] (a00l) rectangle (a00r);
            \draw [draw=none] (a00l) -- node[] {\small\texttt{A0}} (a00r);

            \node (a10l) [above left of = \wordref{1}{3}, xshift=2.3ex, yshift=-2.5ex] {};
            \node (a10r) [above right of = \wordref{1}{5}, xshift=-1.2ex, yshift=0.5ex] {};
            \draw [fill=midnightblue!25, thick, rounded corners=1mm] (a10l) rectangle (a10r);
            \draw [draw=none] (a10l) -- node[] {\small\texttt{A1}} (a10r);

        \end{dependency}
        \caption{Original structure: arguments of the predicate are located in the upper half-plane of the sentence and do not overlap with each other.}
        \label{fig:origin-srl}
    \end{subfigure}
    \hfill
    \begin{subfigure}[b]{0.99\columnwidth}
        \centering
        \begin{dependency}
            \begin{deptext}[column sep=0.2cm,font=\small]
                They\scriptsize$_1$ \& \brickred{\emph{want}}\scriptsize$_2$ \& to\scriptsize$_3$ \& do\scriptsize$_4$ \& more\scriptsize$_5$ \&.\scriptsize$_6$ \\
            \end{deptext}

            \node (a00l) [above left of = \wordref{1}{1}, xshift=0.5ex, yshift=-2.5ex] {};
            \node (a00r) [above right of = \wordref{1}{1}, xshift=-1.2ex, yshift=0.5ex] {};
            \draw [draw=none, fill=brickred!25, thick, rounded corners=1mm] (a00l) rectangle (a00r);

            \node (a10l) [above left of = \wordref{1}{3}, xshift=2.3ex, yshift=-2.5ex] {};
            \node (a10r) [above right of = \wordref{1}{5}, xshift=-1.2ex, yshift=0.5ex] {};
            \draw [draw=none, fill=midnightblue!25, thick, rounded corners=1mm] (a10l) rectangle (a10r);

            \deproot[edge vertical padding=0.6ex, edge height=10ex, label style={fill=black, thick}, edge style={thick}]{2}{\white{\texttt{PRD}}}
            \depedge[edge vertical padding=0.6ex, edge height=4.2ex, label style={fill=brickred!25, thick}, thick, shorten >=2.7ex]{2}{1}{\texttt{A0}}
            \depedge[edge vertical padding=0.6ex, edge height=4.2ex, label style={fill=midnightblue!25, thick}, thick, shorten >=2.7ex]{2}{4}{\texttt{A1}}
            \depedge[edge vertical padding=0.6ex, edge height=6ex, label style={thick}, thick]{2}{6}{$\varnothing$}
            \depedge[<->, edge vertical padding=0.6ex, edge start x offset=-1ex, edge height=2ex, edge slant=2pt, hide label, thick, dotted]{3}{5}{}
            \depedge[<->, edge vertical padding=0.6ex, edge start x offset=0ex, edge end x offset=-0.5ex, edge height=1.5ex, edge slant=4pt, hide label, thick, dotted]{3}{4}{}
            \depedge[<->, edge vertical padding=0.6ex, edge start x offset=-0.5ex, edge end x offset=-1ex, edge height=1.5ex, edge slant=4pt, hide label, thick, dotted]{4}{5}{}

        \end{dependency}
        \caption{Training: convert the predicate-argument structure to a dependency tree with (dotted) latent annotations; non-argument spans are assigned ``$\varnothing$'' for distinction.}
        \label{fig:srl-latent-trees}
    \end{subfigure}
    \hfill
    \begin{subfigure}[b]{0.99\columnwidth}
        \centering
        \begin{dependency}
            \begin{deptext}[column sep=0.2cm,font=\small]
                They\scriptsize$_1$ \& \brickred{\emph{want}}\scriptsize$_2$ \& to\scriptsize$_3$ \& do\scriptsize$_4$ \& more\scriptsize$_5$ \&.\scriptsize$_6$ \\
            \end{deptext}

            \deproot[edge vertical padding=0.6ex, edge height=10ex, label style={fill=black, thick}, edge style={thick}]{2}{\white{\texttt{PRD}}}
            \depedge[edge vertical padding=0.6ex, edge height=4.2ex, label style={fill=brickred!25, thick}, thick]{2}{1}{\texttt{A0}}
            \depedge[edge vertical padding=0.6ex, edge height=4.2ex, label style={fill=midnightblue!25, thick}, thick]{2}{4}{\texttt{A1}}
            \depedge[edge vertical padding=0.6ex, edge height=6ex, label style={thick}, thick, dashed]{2}{6}{$\varnothing$}
            \depedge[edge vertical padding=0.6ex, edge height=2ex, hide label, thick]{4}{3}{}
            \depedge[edge vertical padding=0.6ex, edge height=2ex, hide label, thick]{4}{5}{}

        \end{dependency}
        \caption{Decoding: realize a tree rooted at the predicate with the arc labeled as ``\texttt{PRD}''; (dashed) arcs labeled as ``$\varnothing$'' are discarded.}
        \label{fig:dep-trees}
    \end{subfigure}
    \hfill
    \begin{subfigure}[b]{0.99\columnwidth}
        \centering
        \begin{dependency}
            \begin{deptext}[column sep=0.2cm,font=\small]
                They\scriptsize$_1$ \& \brickred{\emph{want}}\scriptsize$_2$ \& to\scriptsize$_3$ \& do\scriptsize$_4$ \& more\scriptsize$_5$ \&.\scriptsize$_6$ \\
            \end{deptext}

            \node (a00l) [above left of = \wordref{1}{1}, xshift=0.5ex, yshift=-2.5ex] {};
            \node (a00r) [above right of = \wordref{1}{1}, xshift=-1.2ex, yshift=0.5ex] {};
            \draw [fill=brickred!25, thick, rounded corners=1mm] (a00l) rectangle (a00r);

            \node (a10l) [above left of = \wordref{1}{3}, xshift=2.3ex, yshift=-2.5ex] {};
            \node (a10r) [above right of = \wordref{1}{5}, xshift=-1.2ex, yshift=0.5ex] {};
            \draw [fill=midnightblue!25, thick, rounded corners=1mm] (a10l) rectangle (a10r);

            \deproot[edge vertical padding=0.6ex, edge height=10ex, label style={fill=black, thick}, edge style={thick}]{2}{\white{\texttt{PRD}}}
            \depedge[edge vertical padding=0.6ex, edge height=4.2ex, label style={fill=brickred!25, thick}, thick, shorten >=2.7ex]{2}{1}{\texttt{A0}}
            \depedge[edge vertical padding=0.6ex, edge height=4.2ex, label style={fill=midnightblue!25, thick}, thick, shorten >=2.7ex]{2}{4}{\texttt{A1}}
            \depedge[edge vertical padding=0.6ex, edge height=2ex, hide label, thick, dashed]{4}{3}{}
            \depedge[edge vertical padding=0.6ex, edge height=2ex, hide label, thick, dashed]{4}{5}{}
        \end{dependency}
        \caption{Recovery: collapse all (dashed) subtrees governed by the predicate into flat argument spans.}
        \label{fig:collation}
    \end{subfigure}
    \caption{
        Illustration of our SRL$\rightarrow$Tree conversion (Fig.~\ref{fig:origin-srl} and Fig.~\ref{fig:srl-latent-trees}), and its inverse Tree$\rightarrow$SRL process (Fig.~\ref{fig:dep-trees} and Fig.~\ref{fig:collation}).
        We emphasize the predicate ``\brickred{\emph{want}}'' in the figures for clarity.
        The two arguments with roles ``\texttt{A0}'' and ``\texttt{A1}'' are framed by red and blue rectangles, respectively.
    }
    \label{fig:srl-processes}
\end{figure*}

In span-style SRL, an argument of a predicate corresponds to one word or multiple continuous words.
In the latter case, each word in the argument span is treated as equal, and the internal structure of a multi-word argument, i.e., the relationship between words inside the argument, is usually overlooked due to the lack of corresponding annotations.

In this work, we propose to explicitly model internal structures of multi-word arguments and treat arguments as latent subtrees.
Our approach deals with each predicate separately, and assumes each corresponds to a single-root tree.
Consequently, each argument subtree is attached to the predicate.
During the training process, all possible structures are enumerated and accumulated to compose the argument representation.
While decoding, we seek to find a 1-best tree and recover arguments from the subtrees belonging to the resulting structure.
We highlight four key points.
\begin{enumerate}[leftmargin=15pt,label=\roman*]
    \item Our approach is syntax-agnostic.
          The tree structures are modeled and predicted solely to serve the SRL task without referring to any linguistic syntax knowledge.
    \item The predicate identification subtask is handled as a simple classification procedure.
    \item For argument identification, argument boundaries are decided by subtrees attached to the predicate, and edge labels are used for role disambiguation.
    \item We adopt a consistent scoring architecture for the two subtasks and train them jointly.
\end{enumerate}

\subsection{SRL $\rightarrow$ Tree Conversion}\label{sec:srl-tree}

Formally, given an input sentence $\boldsymbol{x}=x_1,\dots, x_n$, we first seek to obtain tree structures for each predicate $p\in \boldsymbol{x}$, which are taken as materials of training a parser.
We define a directed acyclic dependency tree $\boldsymbol{t}$ by assigning a head $h\in\{x_0,x_1,\dots,x_n\}$ together with a relation label $r\in \mathcal{R}$ to each modifier $m \in \boldsymbol{x}$, where a dummy word $x_0$ is attached before $\boldsymbol{x}$ as the pseudo root node.\footnote{
    In this work, we assume all dependency trees are \emph{projective}, i.e., without any crossing arcs.
    This property allows us to associate the subtree with its continuous argument span \cite{kong-etal-2015-transforming}.
}

For predicate $p$, the first step is to link $x_0$ to $p$.
To facilitate predicate identification, we assign a special label \texttt{PRD} (resp. $\varnothing$ for non-predicate) to the dependency $x_0 \rightarrow p$.
Then, we make all corresponding latent argument subtrees descendants of $p$.
As we showcase in Fig.~\ref{fig:origin-srl}, this takes advantage of the non-overlapping constraint for arguments belonging to the same predicate \cite{punyakanok-etal-2004-semantic,li-etal-2019-dependency}.
For an argument with a consecutive word span $x_i,\dots,x_j$ and a semantic role $r\in \mathcal{R}$, we restrict all possible subtrees are single-rooted at a potential headword $h$ within the span, which is also not realized yet.
The semantic role $r$ is assigned as the label of the dependency pointing from $p$ to the headword.
We adopt a similar strategy for non-argument spans, except that we set the label to $\varnothing$  for distinction and remove the single-root restriction.

By enumerating all possible subtrees and combing them together, the resulting tree set $T_p$ is exponential in size.
During training, we develop a span-constrained Inside algorithm to perform the enumeration (\S~\ref{sec:span-constrained-treecrf}).
Fig.~\ref{fig:srl-latent-trees} gives a brief example of the conversion process.

\subsection{Tree $\rightarrow$ SRL Recovery}\label{sec:srl-recovery}
Supposing we have trained a parsing model, during the decoding phase, what we need is to recover predicate-argument structures from the outputs of the parser.

We first find all predicates via simple local label classification: a word $p$ is recognized as a predicate if the dependency $x_0\rightarrow p$ is labeled as \texttt{PRD}.
Subsequently, we obtain the highest-scoring tree $\boldsymbol{t}^{\ast}$ (Fig.~\ref{fig:dep-trees}) for $p$ using Eisner algorithm \cite{eisner-2000-bilexical} with complexity $O(n^3)$:
\begin{equation}%
    \boldsymbol{t}^{\ast} = \arg\max_{\substack{\boldsymbol{t}: x_0 \xrightarrow[]{\texttt{PRD}} p\in\boldsymbol{t}}}
    \mathrm{s}(\boldsymbol{x},\boldsymbol{t})
\end{equation}
where $\mathrm{s}(\boldsymbol{x},\boldsymbol{t})$ is the tree score, and the tree is restricted to be rooted at $p$.
Arguments for the predicate are then recovered by collapsing subtrees headed by $p$ into flat spans.

Concretely, we take each modifier $h$ of $p$ as the headword of a potential argument.
If the label $r$ of $p \rightarrow h$ is not ``$\varnothing$'', i.e., non-argument, then an entire argument span comprises $h$ and its descendants and takes $r$ as the semantic role.
The resulting SRL output is the collection of all predicates and corresponding recovered arguments.
A recovery example is demonstrated in Fig.~\ref{fig:collation}.

\section{Methodology}\label{sec:methodology}

Now we elaborate the architecture of our proposed model for training the parser.
Following \citet{dozat-etal-2017-biaffine,zhang-etal-2020-efficient}, our model consists of a contextualized encoder and a (second-order) scoring module.
We further propose a span-aware TreeCRF to compute the probabilities of the converted partially-observed trees.

\subsection{Neural Parameterization}
Given the sentence $\boldsymbol{x}=x_0,x_1,\dots,x_n$, we first obtain the hidden representation of each token $x_i$ via a deep contextualized encoder.
\begin{equation}
    \mathbf{h}_0,\mathbf{h}_1,\dots,\mathbf{h}_n=\mathtt{Encoder}(x_0,x_1,\dots,x_n)
\end{equation}
In this work, we experiment with two alternative encoders, i.e., BiLSTMs \cite{yarin-etal-2016-dropout} and pretrained language models (PLMs) \cite{devlin-etal-2019-bert}.
More setting details are available in \S~\ref{sec:impl}.

\paragraph{(Second-order) Tree parameterization}
Following \citet{dozat-etal-2017-biaffine}, we decompose a tree $\boldsymbol{t}$ into two separate $\boldsymbol{y}$ and $\boldsymbol{r}$, where $\boldsymbol{y}$ is a skeletal tree, and $\boldsymbol{r}$ is the related strictly-ordered label sequence.
For each head-modifier pair $h\rightarrow m \in \boldsymbol{y}$, we score them using two MLPs followed by a Biaffine layer \cite{cai-etal-2018-full}:
\begin{equation}
    \begin{aligned}
        \mathbf{r}^{\mathrm{head}/\mathrm{mod}}_i & = \mathtt{MLP}^{\mathrm{head}/\mathrm{mod}}(\mathbf{h}_i)                                \\
        \mathrm{s}(h\rightarrow m)                & = \mathtt{BiAF}\left(\mathbf{r}^{\mathrm{head}}_{h},\mathbf{r}^{\mathrm{mod}}_{m}\right)
    \end{aligned}
\end{equation}
The score of the dependency $h\rightarrow m$ with label $r\in \mathcal{R}$ is calculated analogously.
We use two extra MLPs and $|\mathcal{R}|$ Biaffine layers to obtain all label scores.

We also make use of adjacent-sibling information \cite{mcdonald-pereira-2006-online} to enhance the first-order biaffine parser further.
Following \citet{wang-etal-2019-second,zhang-etal-2020-efficient}, we employ three extra MLPs as well as a Triaffine layer for second-order subtree scoring,
\begin{equation}
    \begin{aligned}
        \mathbf{r}^{\mathrm{head}/\mathrm{mod}/\mathrm{sib}}_i & = \mathtt{MLP}^{\mathrm{head}/\mathrm{mod}/\mathrm{sib}}(\mathbf{h}_i)                                                  \\
        \mathrm{s}(h\rightarrow {s,m})                         & = \mathtt{TriAF}\left(\mathbf{r}^{\mathrm{head}}_{h},\mathbf{r}^{\mathrm{mod}}_{m},\mathbf{r}^{\mathrm{sib}}_{s}\right)
    \end{aligned}
\end{equation}
where $s$ and $m$ are two adjacent modifiers of $h$, and $s$ populates between $h$ and $m$.

Under the first-order factorization \cite{mcdonald-etal-2005-online}, the score of $\boldsymbol{y}$ becomes
\begin{equation}\label{eq:tree-score}
    \mathrm{s}(\boldsymbol{x},\boldsymbol{y})=\sum\limits_{h\rightarrow m \in \boldsymbol{y}}\mathrm{s}(h\rightarrow m)
\end{equation}
For the second-order case \cite{mcdonald-pereira-2006-online}, we further incorporate adjacent-sibling subtree scores into tree scoring:
\begin{equation}\label{eq:2otree-score}
    \mathrm{s}(\boldsymbol{x},\boldsymbol{y})=\sum\limits_{h\rightarrow m}\mathrm{s}(h\rightarrow m)+\sum\limits_{h\rightarrow {s,m}}\mathrm{s}(h\rightarrow {s,m})
\end{equation}

The probabilities of skeletal tree $\boldsymbol{y}$ and its label sequence $\boldsymbol{r}$ are parameterized as
\begin{equation}
    \begin{aligned}
        P(\boldsymbol{y}\mid \boldsymbol{x})               & =\frac{\exp\left(\mathrm{s}(\boldsymbol{x},\boldsymbol{y})\right)}{Z(\boldsymbol{x})\equiv\sum_{\boldsymbol{y}^{\prime}\in Y(\boldsymbol{x})}\exp\left(\mathrm{s}(\boldsymbol{x},\boldsymbol{y}^{\prime})\right)} \\
        P(\boldsymbol{r}\mid\boldsymbol{x},\boldsymbol{y}) & =\prod\limits_{h\xrightarrow[]{r}m \in \boldsymbol{t}} P(r\mid \boldsymbol{x},h\rightarrow m)                                                                                                                     \\
    \end{aligned}
\end{equation}
$Y(\boldsymbol{x})$ is the set of all possible legal unlabeled trees, and $Z(\boldsymbol{x})$ is known as the partition function.
Each label $r$ is independent of tree $\boldsymbol{y}$ and other labels, thus $P(r\mid \boldsymbol{x},h\rightarrow m)$ is locally normalized over all $r^{\prime}\in \mathcal{R}$.

Finally, we define the probability of the labeled tree $\boldsymbol{t}$ as the product of the probabilities of its two sub-components.
\begin{equation}
    P(\boldsymbol{t}\mid \boldsymbol{x}) = P(\boldsymbol{y}\mid\boldsymbol{x})\cdot P(\boldsymbol{r}\mid\boldsymbol{x},\boldsymbol{y})
\end{equation}

\subsection{Span-constrained TreeCRF}\label{sec:span-constrained-treecrf}

\paragraph{Training objective}
During training, we seek to maximize the probability of converted trees $T_p$ for each predicate $p$.
Accordingly, we define the following loss function:
\begin{equation}
    \mathcal{L} = - \sum_p\log P(T_p\mid\boldsymbol{x}) \\
\end{equation}
in which $P(T_p\mid\boldsymbol{x})$ can be further expanded as
\begin{equation}\label{eq:prob}
    \begin{aligned}
        P(T_p \mid\boldsymbol{x}) & =\sum_{\boldsymbol{t}\in T_p} \underbrace{P(\boldsymbol{y}\mid \boldsymbol{x})\cdot P(\boldsymbol{r}\mid \boldsymbol{x}, \boldsymbol{y})}_{P(\boldsymbol{t}\mid\boldsymbol{x})}                                                  \\
                                  & = \frac{1}{Z(\boldsymbol{x})}\sum_{\boldsymbol{t}\in T_p}\underbrace{\exp(\mathrm{s}(\boldsymbol{x},\boldsymbol{y}))\cdot P(\boldsymbol{r}\mid \boldsymbol{x},\boldsymbol{y})}_{\exp(\mathrm{s}(\boldsymbol{x},\boldsymbol{t}))}
    \end{aligned}
\end{equation}

\begin{figure}[tb!]
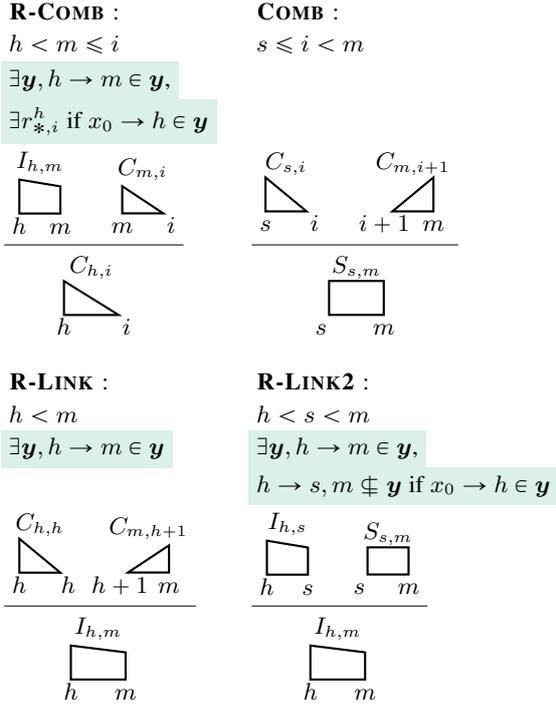

    \renewcommand{\arraystretch}{1}
    \centering
    \small
    \begin{tabular}{ll}
        \textbf{\colorbox{white}{\textsc{R-Comb}}}:\hfill                                                                                                                                                                                                                                                                        & \textbf{\colorbox{white}{\textsc{Comb}}}:    \\\\[-10pt]
        \begin{tabular}[t]{@{}l@{}} \colorbox{white}{$h < m \leq i$}   \\ \colorbox{seagreen!25}{$\exists \boldsymbol{y},h\rightarrow m \in \boldsymbol{y}$,}\\\colorbox{seagreen!25}{$\exists r^h_{*,i}\;\textrm{if}\;x_0\rightarrow h\in \boldsymbol{y}$} \end{tabular} &
        \colorbox{white}{$s \leq i <m$}
        \\[2pt]
        $\inferrule{
                \tikz[baseline=0pt]{\trapezoid[0.6][0.5][0.1]{h}{m}{I_{h,m}}} \quad
                \tikz[baseline=2.5pt]{\righttriangle[0.6][0.4]{m}{i}{C_{m,i}}}
            } {
                \tikz[baseline=0pt]{\righttriangle[0.8][0.5]{h}{i}{C_{h,i}}}
        }$                                                                                                                                                                                                                                                                                                                       &
        $\inferrule{
                \tikz[baseline=0pt]{\righttriangle[0.6][0.5]{s}{i}{C_{s,i}}} \quad
                \tikz[baseline=0pt]{\lefttriangle[0.6][0.5]{i+1}{m}{C_{m,i+1}}}
            } {
                \tikz[baseline=0pt]{\square[0.8][0.5]{s}{m}{S_{s,m}}}
        }$                                                                                                                                                                                                                                                                                                                                                                      \\[40pt]
        \textbf{\colorbox{white}{\textsc{R-Link}}}:                                                                                                                                                                                                                                                                              & \textbf{\colorbox{white}{\textsc{R-Link2}}}: \\\\[-10pt]
        \begin{tabular}[t]{@{}l@{}}\colorbox{white}{$h < m$} \\ \colorbox{seagreen!25}{$\exists \boldsymbol{y}, h\rightarrow m \in \boldsymbol{y}$} \end{tabular}                                                                                                                                      &
        \begin{tabular}[t]{@{}l@{}}\colorbox{white}{$h< s < m$} \\\colorbox{seagreen!25}{$\exists \boldsymbol{y},h\rightarrow m \in \boldsymbol{y}$,}\\\colorbox{seagreen!25}{$h\rightarrow s,m \nsubseteq  \boldsymbol{y}\;\textrm{if}\;x_0\rightarrow h\in \boldsymbol{y}$} \end{tabular}                 \\[2pt]
        $\inferrule{
                \tikz[baseline=0pt]{\righttriangle[0.6][0.5]{h}{h}{C_{h,h}}}
                \tikz[baseline=2.5pt]{\lefttriangle[0.6][0.4]{h+1}{m}{C_{m,h+1}}}
            } {
                \tikz[baseline=0pt]{\trapezoid[0.8][0.5][0.1]{h}{m}{I_{h,m}}}
            }$
                                                                                                                                                                                                                                                                                                                                 &
        $\inferrule{
                \tikz[baseline=0pt]{\trapezoid[0.6][0.5][0.1]{h}{s}{I_{h,s}}} \quad
                \tikz[baseline=2.5pt]{\square[0.6][0.4]{s}{m}{S_{s,m}}}
            } {
                \tikz[baseline=0pt]{\trapezoid[0.8][0.5][0.1]{h}{m}{I_{h,m}}}
            }
        $
    \end{tabular}
    \caption{
    Deduction rules for our span-constrained Inside algorithm (\textbf{\textsc{R-Comb}} and \textbf{\textsc{R-Link}}) and its second-order extension (\textbf{\textsc{Comb}} and \textbf{\textsc{R-Link2}}).
    Our modified rule constraints are highlighted in \emph{green color}.
    The condition $x_0\rightarrow h\in\boldsymbol{y}$ means $h$ is a predicate with $x_0$ as the parent.
    $r^h_{*,i}$ denotes an argument span that takes $h$ as the predicate and ends with $i$.
    We show only R-rules, omitting the symmetric L-rules as well as initial conditions for brevity.
    }
    \label{fig:deduction}
\end{figure}

\paragraph{The proposed Inside}
The calculation of the denominator $Z(\boldsymbol{x})$ in Eq.~\ref{eq:prob} can be accomplished by the canonical Inside algorithm.
As for the numerator, we make a slight change of the formula and define the labeled tree score as:
\begin{equation}
    \mathrm{s}(\boldsymbol{x},\boldsymbol{t})=\mathrm{s}(\boldsymbol{x},\boldsymbol{y})+ \log P(\boldsymbol{r}\mid \boldsymbol{x},\boldsymbol{y})
\end{equation}
In this way, the numerator is exactly the summation of exponential scores of all legal labeled trees.\footnote{
    It is noteworthy that we do not assign any label to $h\rightarrow m \in \boldsymbol{y}$ while $h\notin \{x_0,p\}$, i.e., any dependency inside an argument span, thus its logarithmic label probability is set to 0 and does not contribute to tree scoring.
}
This differs from the traditional case of partial tree learning \cite{li-etal-2016-active} from two perspectives where the common Inside algorithm is not adequate to: 1) we impose span constraints to force the converted latent subtrees to reflect argument spans, and 2) we require the subtree ought to be single-rooted at one potential headword in the span.

To resolve this, in this work, we propose a span-constrained Inside algorithm to accommodate these constraints.
We illustrate the deduction rules \cite{pereira-warren-1983-parsing} of our tailored algorithm and its second-order extension in Fig.~\ref{fig:deduction}.\footnote{We refer interested readers to \S~\ref{sec:inside} for more details on the exact meaning of the operations in the Inside algorithm.}
Basically, we avoid the arc $h\rightarrow m$ crossing different argument spans by prohibiting merging to the relevant incomplete span $I_{h, m}$ (\textbf{\textsc{R-Link}}).
To prevent multiple headwords in the same argument, inspired by \citet{zhang-etal-2021-adapting}, for predicate $h$, we only allow merging to the complete span $C_{h, i}$ if $i$ is at the endpoint of an argument (\textbf{\textsc{R-Comb}}).
For the second-order case, we further prohibit the subtree $h\rightarrow s,m$ once $s$ and $m$ are located in the same argument (\textbf{\textsc{R-Link2}}), since this case implies that the argument can be split into two more smaller headed spans with respect to $s$ and $m$, which is not what we expect.

\paragraph{Time complexity analysis}
The proposed span-constrained Inside shares the same asymptotic time complexity of $O(n^3)$ as its canonical counterpart \cite{eisner-2000-bilexical}.
Besides, we draw on the recent development of parallelization techniques \cite{eisner-2016-inside,zhang-etal-2020-efficient,rush-2020-torch} and further reduce the complexity of the parallelized algorithm to $O(n^2)$ on GPUs.
In practice, we find that our models are efficient enough compared to BIO-based and Span-based models.
We make comprehensive speed comparisons in \S~\ref{sec:speed}.

\section{Experiments}\label{sec:exp}

\begin{table}[tb!]
    \renewcommand{\arraystretch}{1.1}
    \setlength{\tabcolsep}{5pt}
    \centering
    \begin{small}
        \begin{tabular}{l cccc|c}
            \toprule
            \rowcolor[gray]{0.95} & \#Train & \#Dev & \#Test & \#OOD & \#roles \\
            \midrule
            \textsc{CoNLL05}      & 39,832  & 1,346 & 2,416  & 2,399 & 54      \\
            \textsc{CoNLL12}      & 75,187  & 9,603 & 9,479  & -     & 63      \\
            \bottomrule
        \end{tabular}
        \caption{Data statistics for CoNLL05 and CoNLL12.}
        \label{table:statistics}
    \end{small}
\end{table}

We measure our proposed first-order \textsc{Crf} and second-order \textsc{Crf2o} models on two SRL benchmarks: CoNLL05 and CoNLL12.
Full implementation details are given in \S~\ref{sec:impl}.
\paragraph{Data}
Table~\ref{table:statistics} lists the statistics of the datasets.
For CoNLL05, we follow standard splits of \citet{carreras-marquez-2005-introduction}: sections 02-21 of WSJ corpus as Train data, section 24/23 as Dev/Test data, and three sections (CK01-03) of the Brown corpus as out-of-domain (OOD) data.
For CoNLL12, following \citet{he-etal-2018-jointly}, we extract data from OntoNotes \cite{pradhan-etal-2013-towards} and follow the data splits of the CoNLL12 shared task \cite{pradhan-etal-2012-conll}.\footnote{The list of file IDs for Train/Dev/Test data is available on the \href{https://cemantix.org/conll/2012/download/ids/english/coref/}{task webpage}.}
We adopt the same splits for both \emph{end-to-end} and \emph{w/ gold predicates} settings.
We use the official scripts provided by CoNLL05 shared task\footnote{\url{https://www.cs.upc.edu/~srlconll}} for evaluation.

\begin{table*}[tb!]
    \renewcommand{\arraystretch}{1.1}
    \newcolumntype{C}{>{\centering\arraybackslash}m{3.0cm}}
    \newcolumntype{L}{>{\raggedright\arraybackslash}m{3.2cm}}
    \newcolumntype{R}{>{\raggedleft\arraybackslash}m{3.0cm}}
    \setlength{\tabcolsep}{5.9pt}
    \centering
    \begin{small}
        \begin{tabular}{L c ccc ccc c ccc}
            \toprule
            \rowcolor[gray]{0.95}                                             & \multicolumn{7}{c}{CoNLL05} & \multicolumn{4}{c}{CoNLL12}                                                                                                                                                                                       \\
            \rowcolor[gray]{0.95}                                             & Dev                         & \multicolumn{3}{c}{WSJ}     & \multicolumn{3}{c}{Brown} & Dev            & \multicolumn{3}{c}{Test}                                                                                                               \\
            \rulefiller\cmidrule(lr){2-2}	\cmidrule(lr){3-5}		\cmidrule(lr){6-8} \cmidrule(lr){9-9}	\cmidrule(lr){10-12}
            \rowcolor[gray]{0.95}                                             & F$_1$                       & P                           & R                         & F$_1$          & P                        & R              & F$_1$          & F$_1$          & P                      & R              & F$_1$          \\
            \midrule
            \citet{he-etal-2017-deep}                                         & 80.3\white{0}               & 80.2\white{0}               & 82.3\white{0}             & 81.2\white{0}  & 67.6\white{0}            & 69.6\white{0}  & 68.5\white{0}  & 75.5\white{0}  & 78.6\white{0}          & 75.1\white{0}  & 76.8\white{0}  \\
            \citet{he-etal-2018-jointly}                                      & 81.6\white{0}               & 81.2\white{0}               & 83.9\white{0}             & 82.5\white{0}  & 69.7\white{0}            & 71.9\white{0}  & 70.8\white{0}  & 79.4\white{0}  & 79.4\white{0}          & 80.1\white{0}  & 79.8\white{0}  \\
            \citet{li-etal-2019-dependency}                                   & -                           & -                           & -                         & 83.0\white{0}  & -                        & -              & -              & -              & -                      & -              & -              \\
            \citet{zhou-etal-2020-parsing}                                    & 82.27                       & -                           & -                         & -              & -                        & -              & -              & -              & -                      & -              & -              \\\\[-10pt]
            \textsc{Crf}                                                      & 83.70                       & 83.18                       & 85.38                     & 84.27          & 70.40                    & 72.97          & 71.66          & 81.03          & \textbf{79.47}         & 82.80          & 81.10          \\
            \textsc{Crf2o}                                                    & \textbf{83.91}              & \textbf{83.26}              & \textbf{86.20}            & \textbf{84.71} & \textbf{70.70}           & \textbf{74.16} & \textbf{72.39} & \textbf{81.16} & 79.27                  & \textbf{83.24} & \textbf{81.21} \\\\[-10pt]
            \citet{li-etal-2019-dependency}$_\texttt{ELMo}$                   & -                           & 85.2\white{0}               & 87.5\white{0}             & 86.3\white{0}  & 74.7\white{0}            & 78.1\white{0}  & 76.4\white{0}  & -              & \textbf{84.9\white{0}} & 81.4\white{0}  & 83.1\white{0}  \\
            \citet{zhou-etal-2022-fast}\rlap{\rlap{$_\texttt{BERT}$}}         & 86.79                       & \textbf{87.15}              & 88.44                     & 87.79          & \textbf{79.44}           & 80.85          & 80.14          & 84.74          & 83.91                  & 85.61          & 84.75          \\
            \textsc{Crf}\rlap{$_\texttt{BERT}$}                               & 86.82                       & 86.98                       & 88.28                     & 87.63          & 79.19                    & 80.92          & 80.05          & 85.35          & 84.47                  & 86.24          & 85.35          \\
            \textsc{Crf2o}\rlap{$_\texttt{BERT}$}                             & \textbf{87.03}              & 87.00                       & \textbf{88.76}            & \textbf{87.87} & 79.08                    & \textbf{81.50} & \textbf{80.27} & \textbf{85.53} & 84.53                  & \textbf{86.41} & \textbf{85.45} \\
            \hdashline[1pt/3pt]
            \textsc{Crf}\rlap{$_\texttt{RoBERTa}$}                            & 87.31                       & 87.20                       & 88.67                     & 87.93          & 79.29                    & 81.48          & 80.38          & 86.08          & 84.98                  & 86.86          & 85.91          \\
            \textsc{Crf2o}\rlap{$_\texttt{RoBERTa}$}                          & \textbf{87.46}              & \textbf{87.35}              & \textbf{89.34}            & \textbf{88.33} & \textbf{79.95}           & \textbf{82.32} & \textbf{81.12} & \textbf{86.34} & \textbf{85.30}         & \textbf{87.02} & \textbf{86.15} \\
            \rowcolor[gray]{0.95}\multicolumn{12}{c}{\emph{w/ gold predicates}}                                                                                                                                                                                                                                                 \\
            \citet{he-etal-2017-deep}                                         & 81.6\white{0}               & 83.1\white{0}               & 83.0\white{0}             & 83.1\white{0}  & 72.9\white{0}            & 71.4\white{0}  & 72.1\white{0}  & 81.5\white{0}  & 81.7\white{0}          & 81.6\white{0}  & 81.7\white{0}  \\
            \citet{ouchi-etal-2018-span}                                      & 82.5\white{0}               & 84.7\white{0}               & 82.3\white{0}             & 83.5\white{0}  & \textbf{76.0\white{0}}   & 70.4\white{0}  & 73.1\white{0}  & 82.9\white{0}  & \textbf{84.4\white{0}} & 81.7\white{0}  & 83.0\white{0}  \\
            \citet{tan-etal-2018-deep}                                        & 83.1\white{0}               & 84.5\white{0}               & 85.2\white{0}             & 84.8\white{0}  & 73.5\white{0}            & 74.6\white{0}  & 74.1\white{0}  & 82.9\white{0}  & 81.9\white{0}          & 83.6\white{0}  & 82.7\white{0}  \\
            \citet{strubell-etal-2018-lisa}                                   & -                           & 84.7\white{0}               & 84.24                     & 84.47          & 73.89                    & 72.39          & 73.13          & -              & -                      & -              & -              \\
            \citet{zhou-etal-2020-parsing}                                    & 83.16                       & -                           & -                         & -              & -                        & -              & -              & -              & -                      & -              & -              \\
            \citet{zhang-etal-2021-comparing}                                 & 84.45                       & 85.30                       & 85.17                     & 85.23          & 74.98                    & 73.85          & 74.41          & 82.83          & 83.09                  & 83.71          & 83.40          \\\\[-10pt]
            \textsc{Crf}                                                      & 84.42                       & 85.38                       & 85.56                     & 85.47          & 75.05                    & 74.05          & 74.55          & 83.22          & 83.21                  & 83.85          & 83.53          \\
            \textsc{Crf2o}                                                    & \textbf{84.65}              & \textbf{85.47}              & \textbf{86.40}            & \textbf{85.93} & 74.92                    & \textbf{75.00} & \textbf{74.96} & \textbf{83.39} & 83.02                  & \textbf{84.31} & \textbf{83.66} \\\\[-10pt]
            \citet{strubell-etal-2018-lisa}\rlap{$_\texttt{ELMo}$}            & 85.26                       & 86.21                       & 85.98                     & 86.09          & 77.1\white{0}            & 75.61          & 76.35          & 83.23          & 84.39                  & 82.21          & 83.28          \\
            \citet{shi-etal-2019-simple}\rlap{$_\texttt{BERT}$}               & -                           & 88.6\white{0}               & 89.0\white{0}             & 88.8\white{0}  & 81.9\white{0}            & 82.1\white{0}  & 82.0\white{0}  & -              & 85.9\white{0}          & 87.0\white{0}  & 86.5\white{0}  \\
            \citet{jindal-etal-2020-improved}\rlap{$_\texttt{BERT}$}          & -                           & 88.7\white{0}               & 88.0\white{0}             & 87.9\white{0}  & 80.3\white{0}            & 80.1\white{0}  & 80.2\white{0}  & -              & 86.3\white{0}          & 86.8\white{0}  & 86.6\white{0}  \\
            \citet{zhang-etal-2021-comparing}\rlap{\rlap{$_\texttt{BERT}$}}   & 87.38                       & 87.70                       & 88.15                     & 87.93          & 81.52                    & 81.36          & 81.44          & 86.27          & 86.00                  & 86.84          & 86.42          \\
            \citet{zhou-etal-2022-fast}\rlap{\rlap{$_\texttt{BERT}$}}         & 87.54                       & \textbf{89.03}              & 88.53                     & 88.78          & \textbf{83.22}           & 81.81          & 82.51          & 86.97          & 87.26                  & 87.05          & 87.15          \\
            \citet{conia-navigli-2020-bridging}\rlap{\rlap{$_\texttt{BERT}$}} & -                           & -                           & -                         & -              & -                        & -              & -              & -              & 86.9\white{0}          & 87.7\white{0}  & 87.3\white{0}  \\
            \citet{blloshmi-etal-2021-generating}\rlap{$_\texttt{BART}$}      & -                           & -                           & -                         & -              & -                        & -              & -              & -              & \textbf{87.8}\white{0} & 86.8\white{0}  & 87.3\white{0}  \\\\[-10pt]
            \textsc{Crf}\rlap{$_\texttt{BERT}$}                               & 87.76                       & 88.93                       & 88.58                     & 88.76          & 82.87                    & 81.67          & 82.27          & 87.33          & 87.45                  & 87.56          & 87.51          \\
            \textsc{Crf2o}\rlap{$_\texttt{BERT}$}                             & \textbf{88.05}              & 89.00                       & \textbf{89.03}            & \textbf{89.02} & 82.81                    & \textbf{82.35} & \textbf{82.58} & \textbf{87.52} & 87.52                  & \textbf{87.79} & \textbf{87.66} \\
            \hdashline[1pt/3pt]
            \textsc{Crf}\rlap{$_\texttt{RoBERTa}$}                            & 88.21                       & 89.29                       & 88.99                     & 89.15          & 83.22                    & 82.42          & 82.82          & 87.97          & 87.99                  & 88.22          & 88.11          \\
            \textsc{Crf2o}\rlap{$_\texttt{RoBERTa}$}                          & \textbf{88.49}              & \textbf{89.45}              & \textbf{89.63}            & \textbf{89.54} & \textbf{83.89}           & \textbf{83.39} & \textbf{83.64} & \textbf{88.29} & \textbf{88.11}         & \textbf{88.53} & \textbf{88.32} \\
            \bottomrule
        \end{tabular}
        \caption{
            All results on CoNLL05 and CoNLL12 data, averaged over 4 runs with different random seeds.
        }
        \label{table:main-results}
    \end{small}
\end{table*}

\subsection{Main results}\label{subsec:results}
Table~\ref{table:main-results} gives our main results.
By default, our models work in an \emph{end-to-end} fashion, i.e., predicting all predicates and their associated arguments simultaneously.
However, we note that reporting the results of using gold predicates is a more prevalent practice in the SRL community \cite{he-etal-2018-jointly,shi-etal-2019-simple}.
Therefore, for comprehensive comparisons, in addition to listing most \emph{end-to-end} results of previous works we are aware of, we also conduct experiments with gold predicates, which is achieved by only parsing trees rooted at the pre-specified predicates.\footnote{We eliminate the invalid $x_0\rightarrow p$ simply via setting the dependency score to $-\infty$.}

The two major rows show the results of \emph{end-to-end} and \emph{w/ gold predicates} settings, indicating very consistent trends.
We can clearly see that under the \emph{end-to-end} setting, our LSTM-based \textsc{Crf} models outperform previous works by a large margin on all datasets.
The second-order \textsc{Crf2o} further improves over \textsc{Crf} by 0.2, 0.4 and 0.7 F$_1$ scores on three CoNLL05 datasets, respectively.
On CoNLL12, \textsc{Crf2o} shows smaller but steady gains.
As revealed in \S~\ref{subsec:analysis}, we attribute the improvements brought by \textsc{Crf} and \textsc{Crf2o} to better performing at global consistency and long-range dependencies.

The results under the \emph{w/ gold predicates} setting are presented in the second major row.
Many PLM-based results comparable to ours are available in this setting.
Among them, the BIO-based parser of \citet{shi-etal-2019-simple} achieves 88.8, 82.0 and 86.5 F$_1$ scores on CoNLL05 WSJ, Brown and CoNLL12 Test data.
The dependency (word)-based parser of \citet{zhou-etal-2022-fast} achieves 88.78,  82.51 and 87.15 F$_1$ scores.
Meanwhile, the results of our first-order \textsc{Crf} model with BERT is 88.76$_{\pm 0.18}$, 82.27$_{\pm 0.26}$ and 87.51$_{\pm 0.11}$.
The performance gap between \textsc{Crf} and recent state-of-the-art parsers are negligibly small.
We do not utilize any word/predicate embeddings as well as LSTM layers for simplicity, which may potentially hinder the results.
Despite this fact, our second-order \textsc{Crf2o} achieves 89.02$_{\pm 0.17}$, 82.58$_{\pm 0.47}$, and 87.66$_{\pm 0.05}$, which outperforms the systems of \citet{shi-etal-2019-simple} by 0.2, 0.6 and 1.2 F$_1$ scores and achieves new state-of-the-art on both CoNLL05 and CoNLL12 datasets.
This implies that imposing stronger structure constraints can still bring remarkable improvements for span-style SRL even when empowered with very expressive encoders.
In the bottom lines, we provide the results of utilizing RoBERTa, we can see that \textsc{Crf} and \textsc{Crf2o} augmented with RoBERTa can obtain further gains on top of BERT.

We highlight that we do not include any syntax-aware work \cite{xia-etal-2019-syntax,zhou-etal-2020-parsing} in Table~\ref{table:main-results}, which has shown to deliver substantial gains for SRL (see \S~\ref{sec:incomparable}).
It is still an open question to be investigated whether the benefits brought by our methods are orthogonal to linguistic syntax knowledge.
We focus on pure syntax-agnostic models in this paper.
So we do not list the results of this line of works in order to make fair comparisons.

\subsection{Analysis}\label{subsec:analysis}

\begin{table}[tb!]
    \renewcommand{\arraystretch}{1.1}
    \setlength{\tabcolsep}{4.5pt}
    \centering
    \begin{small}
        \begin{tabular}{l cccccc}
            \toprule
            \rowcolor[gray]{0.95} & \multicolumn{4}{c}{Dev} & \multicolumn{2}{c}{Test}                                                                     \\
            \rulefiller\cmidrule(lr){2-5} \cmidrule(lr){6-7}
            \rowcolor[gray]{0.95} & P                       & R                        & F$_1$          & CM             & F$_1$          & CM             \\
            \midrule
            \textsc{Bio}          & 86.80                   & 86.38                    & 86.59          & 69.24          & 88.22          & 71.95          \\
            \textsc{Span}         & 87.68                   & 86.75                    & 87.21          & 68.43          & 88.44          & 70.22          \\\\[-10pt]
            \textsc{Crf}          & 87.89                   & 87.62                    & 87.76          & 71.59          & 88.76          & 73.01          \\
            \quad\textsc{First}   & 87.44                   & 86.60                    & 87.02          & 70.35          & 87.81          & 71.14          \\
            \quad\textsc{Last}    & 86.99                   & 87.00                    & 86.99          & 70.29          & 87.67          & 71.08          \\
            \quad\textsc{Flat}    & 85.63                   & 82.26                    & 83.91          & 63.41          & 83.32          & 62.22          \\
            \textsc{Crf2o}        & \textbf{88.02}          & \textbf{88.09}           & \textbf{88.05} & \textbf{72.57} & \textbf{89.02} & \textbf{73.74} \\
            \bottomrule
        \end{tabular}
        \caption{Finetuning results on CoNLL05 Dev and Test data under the setting of \emph{w/ gold predicates}.}
        \label{table:conll05-dev}
    \end{small}
\end{table}

To better understand which empowers our proposed \textsc{Crf} and \textsc{Crf2o} and in what aspects they are helpful, we conduct detailed analyses on CoNLL05 Dev data.
Considering that there exist many differences in model/training settings, we re-implemented the following two methods based on two widely used libraries HanLP\footnote{\url{https://github.com/hankcs/HanLP}} \cite{he-choi-2021-stem} and SuPar\footnote{\url{https://github.com/yzhangcs/parser}} for fair comparisons:
\begin{itemize}[leftmargin=11pt]
    \item \textbf{\textsc{Bio}}: BIO-based method of \citet{zhou-xu-2015-end}.
          Following \citet{zhang-etal-2021-comparing}, we employ linear-chain CRF \cite{lafferty-etal-2001-crf} to conduct global inference during training.
    \item \textbf{\textsc{Span}}: span-based method of \citet{he-etal-2018-jointly}.
          We borrow the settings of \citet{strubell-etal-2018-lisa} and make use of Biaffine layers for span scoring.
\end{itemize}
We adopt the same experimental setups for all implementations, i.e., finetuning on BERT and assuming all predicates are given.
Results are shown in Table~\ref{table:conll05-dev}.
It is clear that under the same settings, our \textsc{Crf} expands the advantages over \textsc{Bio} and \textsc{Span}, and \textsc{Crf2o} further improves the performance.

\paragraph{Impact of latent subtrees}
First of all, we consider three variants of our first-order \textsc{Crf} to verify the necessity of modeling arguments as latent trees:
1) \textsc{First}, similar to \textsc{Crf} but always takes the first word as argument headword;
2) \textsc{Last}, denoting the last word accordingly;
3) \textsc{Flat}, similar to \textsc{First} but directly attach other argument words to the first word.
The first two variants fix the position of argument headwords.
In Table~\ref{table:conll05-dev}, we observe that \textsc{First} and \textsc{Last} perform quite similarly and steadily inferior to \textsc{Crf}.
This agrees with \citet{zhang-etal-2021-comparing}, highlighting the importance of headwords in recognizing arguments.
In contrast to \textsc{Crf}, \textsc{Flat} completely excludes latent representations during training and restricts the height of the converted trees to 2.
We can see that \textsc{Flat} achieves 83.91 F$_1$ on Dev, a dramatic performance drop against \textsc{Crf} (87.76).
Overall, as we expect, it seems that totally latent argument representations empower \textsc{Crf} a lot, performing best compared to other variants.

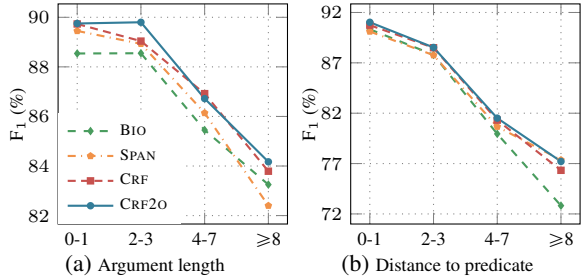
\begin{figure}[tb!]
    \centering
    \begin{small}
        \captionsetup[subfigure]{aboveskip=-1pt,belowskip=-1pt}
        \begin{subfigure}[t]{0.49\columnwidth}
            \centering
            \begin{tikzpicture}
                \begin{axis}[
                        legend style={
                                draw=none,
                                fill opacity=2,
                                text=black,
                                at={(0,0)},
                                anchor=south west,
                                font=\tiny,
                            },
                        symbolic x coords={0-1,2-3,4-7,$\geq$8},
                        xtick=data,
                        font=\scriptsize,
                        ylabel={F$_1$ (\%)},
                        ylabel style = {yshift=-20pt},
                        legend cell align={left},
                        height=4.5cm,
                        width=4.6cm,
                        grid=major,
                        grid style={dotted, color=gray},
                    ]
                    \addplot[mark=diamond*, mark options={solid,mark size=1pt}, dashed,thick,forestgreen] coordinates {
                            (0-1,     88.54)
                            (2-3,     88.55)
                            (4-7,     85.45)
                            ($\geq$8, 83.25)
                        };
                    \addlegendentry{\textsc{Bio}}
                    \addplot[mark=pentagon*, mark options={solid,mark size=1pt}, dashdotted,thick,burntorange] coordinates {
                            (0-1,     89.45)
                            (2-3,     88.93)
                            (4-7,     86.14)
                            ($\geq$8, 82.40)
                        };
                    \addlegendentry{\textsc{Span}}
                    \addplot[mark=square*, mark options={solid,mark size=1pt}, densely dashed,thick,brickred!80] coordinates {
                            (0-1,     89.73)
                            (2-3,     89.04)
                            (4-7,     86.92)
                            ($\geq$8, 83.79)
                        };
                    \addlegendentry{\textsc{Crf}}
                    \addplot[mark=*, mark options={solid,mark size=1pt}, thick,midnightblue!80] coordinates {
                            (0-1,     89.75)
                            (2-3,     89.80)
                            (4-7,     86.72)
                            ($\geq$8, 84.17)
                        };
                    \addlegendentry{\textsc{Crf2o}}
                \end{axis}
            \end{tikzpicture}
            \caption{\scriptsize Argument length}
            \label{fig:f1-arg-len}
        \end{subfigure}
        \hfill
        \begin{subfigure}[t]{0.49\columnwidth}
            \centering
            \begin{tikzpicture}
                \begin{axis}[
                        font=\scriptsize,
                        symbolic x coords={0-1,2-3,4-7,$\geq$8},
                        xtick=data,
                        ytick = {72, 77,..., 92},
                        ylabel={F$_1$ (\%)},
                        ylabel style = {yshift=-20pt},
                        legend cell align={left},
                        height=4.5cm,
                        width=4.6cm,
                        grid=major,
                        grid style={dotted, color=gray},
                    ]
                    \addplot[mark=diamond*, mark options={solid,mark size=1pt}, dashed,thick,forestgreen] coordinates {
                            (0-1,     90.33)
                            (2-3,     87.80)
                            (4-7,     79.95)
                            ($\geq$8, 72.82)
                        };
                    \addplot[mark=pentagon*, mark options={solid,mark size=1pt}, dashdotted,thick,burntorange] coordinates {
                            (0-1,     90.11)
                            (2-3,     87.76)
                            (4-7,     80.66)
                            ($\geq$8, 77.36)
                        };
                    \addplot[mark=square*, mark options={solid,mark size=1pt}, densely dashed,thick,brickred!80] coordinates {
                            (0-1,     90.74)
                            (2-3,     88.50)
                            (4-7,     81.31)
                            ($\geq$8, 76.33)
                        };
                    \addplot[mark=*, mark options={solid,mark size=1pt}, thick,midnightblue!80] coordinates {
                            (0-1,     91.03)
                            (2-3,     88.53)
                            (4-7,     81.51)
                            ($\geq$8, 77.21)
                        };
                \end{axis}
            \end{tikzpicture}
            \caption{\scriptsize Distance to predicate}
            \label{fig:f1-dist}
        \end{subfigure}
    \end{small}
    \caption{
        F$_1$ scores breakdown by argument length (Fig.~\ref{fig:f1-arg-len}) and predicate-argument distance (Fig.~\ref{fig:f1-dist}).
    }
    \label{fig:f1-breakdown}
\end{figure}

\paragraph{Structural consistency}
To quantify the benefits of our methods in making global decisions for SRL structures, we report the percentage of completely correct predicates (CM) \cite{he-etal-2018-jointly} in Table~\ref{table:conll05-dev}.
We show that \textsc{Bio} with linear-chain CRF significantly outperforms \textsc{Span}, but still falls short of our \textsc{Crf} by 1.5.
By explicitly modeling sibling information, \textsc{Crf2o} provides stronger structure constraints and goes further beyond \textsc{Crf} by 0.9.
In terms of the performance broken down by argument length, as shown in Fig.~\ref{fig:f1-arg-len}, \textsc{Span} lags largely behind \textsc{Bio} over length$\geq$8.
We guess this is mainly because of their aggressive argument pruning strategy.
And as expected, \textsc{Crf} and \textsc{Crf2o} demonstrate steady improvements over \textsc{Bio} and \textsc{Span}.
We owe this to the superiority of our formulations in modeling subtree structures, thus providing more powerful argument representations and rich inter- and intra-argument dependency interactions.

\paragraph{Long-range dependencies}
Fig.~\ref{fig:f1-dist} shows the results broken down by predicate-argument distance.
It is clear that the gaps between \textsc{Bio} and other methods become larger as the distance increases.
This is reasonable since \textsc{Bio} lacks explicit connections for non-adjacent predicate-argument pairs, whereas ours provides direct word-to-word bilexical mappings.
\textsc{Span} shows competitive results but is still inferior to ours.
we speculate this is due to their inferiority in ultra-long arguments (see Fig.~\ref{fig:f1-arg-len}).

\subsection{Efficiency}\label{sec:speed}

Table~\ref{table:speed} compares different models in terms of parsing speed.
We obtain the speed of previous works by rerunning their released code.
For fair comparisons, all models are run on Intel Xeon E5-2650 v4 CPU and Nvidia GeForce GTX 1080 Ti GPU, and do not use any PLM layers.

We can clearly see that our \textsc{Crf} and \textsc{Crf2o} can parse about 242 and 214 sentences per second respectively, much faster than all previous works.
In line with \citet{strubell-etal-2018-lisa}, our \textsc{Crf} and \textsc{Crf2o} consume 5,000 tokens (roughly 200 sentences) per mini-batch.
However, \citet{strubell-etal-2018-lisa} use up to 12 Transformer layers, much deeper than our 3-layer BiLSTM encoder.
This explains their less efficiency from the side, as encoder layers might take up a major part of the running time, while the relative more efficient Viterbi decoding does not dominate the time-consuming.
Moreover, our models are based on highly parallelized implementations \cite{zhang-etal-2020-efficient}.
We speculate that the model speed of \citet{strubell-etal-2018-lisa} can be further improved with dedicated optimization.
As for \citet{he-etal-2018-jointly,li-etal-2019-dependency}, we adopt their default setting of 40 sentences per batch.
They need to obtain the representations of all candidate argument spans, leading to high GPU memory usage.
This limits us to enlarge the batch size further and significantly slows down the parsing speed.
In the bottom lines of Table~\ref{table:speed}, we can see that our \textsc{Crf} and \textsc{Crf2o} enhanced with BERT achieve speeds of 136 and 113 sentences per second respectively.
Overall, we can conclude that our proposed \textsc{Crf} and \textsc{Crf2o} are efficient enough and readily applicable to real-life systems.

\begin{table}[t]
    \newcolumntype{L}{>{\raggedright\arraybackslash}m{2.7cm}}
    \centering
    \begin{small}
        \begin{tabular}{lLr}
            \toprule
            \rowcolor[gray]{0.95}           &                                & Sents/s      \\
            \midrule
            \citet{strubell-etal-2018-lisa} & \textsc{Bio}                   & 50           \\
            \citet{he-etal-2018-jointly}    & \textsc{Span}                  & 49           \\
            \citet{li-etal-2019-dependency} & \textsc{Span}                  & 20           \\\\[-10pt]
            \multirow{4}{*}{Ours}           & \textsc{Crf}                   & \textbf{242} \\
                                            & \textsc{Crf2o}                 & 214          \\
                                            & \textsc{Crf}$_\texttt{BERT}$   & 136          \\
                                            & \textsc{Crf2o}$_\texttt{BERT}$ & 113          \\
            \bottomrule
        \end{tabular}
    \end{small}
    \caption{
        Speed comparison on CoNLL05 Test data.
        We also list the speed of our TreeCRF models using BERT (\textsc{Crf}$_\texttt{BERT}$ and \textsc{Crf2o}$_\texttt{BERT}$).
    }
    \label{table:speed}
\end{table}

\section{Related Works}\label{sec:relworks}

\paragraph{Span-style SRL}
Pioneered by \citet{gildea-jurafsky-2002-automatic}, syntax has long been considered indispensable for span-style SRL \cite{punyakanok-etal-2008-importance}.
With the advent of the neural network era, syntax-agnostic models make remarkable progress \cite{zhou-xu-2015-end,tan-etal-2018-deep,cai-etal-2018-full}, mainly owing to powerful model architectures like BiLSTM \cite{yarin-etal-2016-dropout} or Transformer \cite{vaswani-2017-attention}.
Meanwhile, other researchers also pay attention to the utilization of syntax trees, including serving as guidance for argument pruning \cite{he-etal-2018-syntax}, as input features \cite{marcheggiani-titov-2017-encoding,xia-etal-2019-syntax,mohammadshahi-etal-2021-g2g}, or as supervisions for joint learning \cite{swayamdipta-etal-2018-syntactic}.
However, to our best knowledge, very few works have been devoted to mining internal structures of shallow SRL representations.
As exceptions, \citet{he-etal-2018-jointly,zhang-etal-2021-comparing} take into account headwords while recognizing arguments.
Beyond this, this work proposes to model full argument subtree structures rather than merely headwords and find more competitive results.

\paragraph{Parsing with latent variables}
\citet{henderson-etal-2008-latent, henderson-etal-2013-multilingual} design a latent variable model to deliver syntactic and semantic interactions under the setting of joint learning.
In more common situations where gold treebanks may lack, %
\citet{naradowsky-etal-2012-improving,gormley-etal-2014-low} use LBP for the inference of semantic graphs and treat latent trees as global factors %
\cite{smith-eisner-2008-dependency} to provide soft beliefs for reasonable predicate-arguments structures.
This work differs in that we make hard constraints on syntax tree structures to conform to the SRL structures, and take only subtrees attached to predicates as latent variables.
The intuition behind latent tree models \cite{marina-michael-2000-mixure,chu-etal-2017-latent,kim-2017-structured} is to utilize tree structures to provide rich structural interactions for problems with prohibitive high complexity.
This idea is also common in many other NLP tasks like text summarization \cite{liu-lapata-2018-learning}, sequence labeling \cite{zhou-etal-2020-latent}, and AMR parsing \cite{zhou-etal-2020-amr}.

\paragraph{Reduction to tree parsing}
Researchers have investigated several ways to recover SRL structures from tree structures, due to their high coupling nature \cite{palmer-etal-2005-propbank}.
Early efforts of \citet{cohn-blunsom-2005-semantic} derive predicate-arguments from pruned phrase structures by using a CKY-style TreeCRF to learn parameters.
\citet{johansson-nugues-2008-dependency} and \citet{choi-palmer-2010-retrieving} investigate retrieving semantic boundaries from dependency outputs.
Their devised heuristics rely heavily on the quality of output trees, leading to inferior results.
Our reduction is also inspired by works on other NLP tasks, including named entity recognition (NER) \cite{yu-etal-2020-named}, nested NER \cite{fu-etal-2021-nested,lou-etal-2022-nested}, semantic parsing \cite{sun-etal-2017-semantic,jiang-etal-2019-hlt}, and EUD parsing \cite{anderson-gomez-rodriguez-2021-splitting}.
As the most relevant work, \citet{shi-etal-2020-semantic} also propose to reduce SRL to syntactic dependency parsing by integrating syntactic-semantic relations into a single dependency tree by means of joint labels.
However, their approach shows non-improved results, possibly due to the label sparsity problem and high back-and-forth conversion loss.
Also, they use gold treebank supervisions, while ours does not rely on any hand-annotated syntax data.

\section{Discussions and Future Works}
The basic idea of this work is to mimic SRL structures with a combination of multiple latent trees.
This new perspective sheds light on some natural extensions of our work to other tightly related semantic parsing tasks, e.g., AMR \cite{zhang-etal-2019-amr} and UCCA \cite{jiang-etal-2019-hlt}.\footnote{We thank an anonymous reviewer for pointing out the connection.}
Tasks fall into this type exhibit very flexible graph representation schemes (e.g., \emph{reentrancy} and \emph{discontinuity}) \cite{zhang-etal-2019-broad}, which are intractable by principled decoding algorithms like dynamic programming.
We believe that employing structured inference in spirit of our approaches can provide considerable help in getting rid of greedy span/dependency selections and finding globally optimal structures.

We prefer to reduce SRL to dependency-based tree parsing rather than another paradigm, i.e., constituency parsing, partly because dependencies provide a more transparent bilexical governor-dependent encoding of predicate-argument relations \cite{hacioglu-2004-semantic}.
We also do not pursue the way of jointly modeling dependencies and phrasal structures with lexicalized trees \cite{eisner-satta-1999-efficient,yang-tu-2022-combining,lou-etal-2022-nested} as our approach enjoys a lower time complexity of $O(n^3)$.
Nonetheless, we admit potential advantages of this kind of modeling \cite{liu-etal-2022-structured} and leave this as our future work.

There are other interesting perspectives deserve further explorations: given that span-style SRL substantially benefits from our formulation of recovering SRL structures from trees, \emph{can the induced dependency trees learn plausible syntactic structures?} Or in other words, can they agree with linguistic-motivated annotations \cite{marcus-etal-1993-building}?
We conduct thorough analyses in spirit of \citet{gormley-etal-2014-low,li-etal-2021-syntax} and give affirmative answers.
Due to space limitations, we refer readers to \S~\ref{sec:tree-probing} and \S~\ref{sec:induction} for details.

\section{Conclusions}\label{sec:conclusions}

In this paper, we propose to reduce span-style SRL to dependency parsing by viewing flat phrasal arguments as latent subtrees, and design a novel span-constrained TreeCRF to accommodate the span structures.
Taking inspirations from the parsing literature, we also build a second-order extension and find further gains.
Our models are syntax-agnostic and do not rely on any external linguistic syntax knowledge.
Experimental results show that, our proposed methods outperform all previous comparable works, achieving new state-of-the-art on both CoNLL05 and CoNLL12 benchmarks.
Extensive analyses confirm that our approach enjoys some merits of global structural constraints, meanwhile maintaining acceptable time complexity.
Furthermore, we find our modeling of latent subtrees provides effective assistance in terms of long-range dependencies and global consistency.

\section{Acknowledgments}\label{sec:ack}
We would like to thank the anonymous reviewers for their valuable comments, and Prof. Zhenghua Li for very helpful feedbacks, suggestions and idea discussions.
This work was supported by National Natural Science Foundation of China with Grant No. 62176173, 62076173, U1836222 and a project funded by the Priority Academic Program Development (PAPD) of Jiangsu Higher Education Institutions.

\bibliographystyle{acl_natbib}
\bibliography{main}

\begin{thebibliography}{107}
\expandafter\ifx\csname natexlab\endcsname\relax\def\natexlab#1{#1}\fi

\bibitem[{Anderson and
  G{\'o}mez-Rodr{\'\i}guez(2021)}]{anderson-gomez-rodriguez-2021-splitting}
Mark Anderson and Carlos G{\'o}mez-Rodr{\'\i}guez. 2021.
\newblock \href {https://doi.org/10.18653/v1/2021.iwpt-1.17} {Splitting {EUD}
  graphs into trees: A quick and clatty approach}.
\newblock In \emph{Proceedings of IWPT}, pages 167--174, Online.

\bibitem[{Bazrafshan and Gildea(2013)}]{bazrafshan-gildea-2013-semantic}
Marzieh Bazrafshan and Daniel Gildea. 2013.
\newblock \href {https://aclanthology.org/P13-2074} {Semantic roles for string
  to tree machine translation}.
\newblock In \emph{Proceedings of ACL}, pages 419--423, Sofia, Bulgaria.

\bibitem[{Berant et~al.(2013)Berant, Chou, Frostig, and
  Liang}]{berant-etal-2013-semantic}
Jonathan Berant, Andrew Chou, Roy Frostig, and Percy Liang. 2013.
\newblock \href {https://aclanthology.org/D13-1160} {Semantic parsing on
  {F}reebase from question-answer pairs}.
\newblock In \emph{Proceedings of EMNLP}, pages 1533--1544, Seattle,
  Washington, USA.

\bibitem[{Blloshmi et~al.(2021)Blloshmi, Conia, Tripodi, and
  Navigli}]{blloshmi-etal-2021-generating}
Rexhina Blloshmi, Simone Conia, Rocco Tripodi, and Roberto Navigli. 2021.
\newblock \href {https://www.ijcai.org/proceedings/2021/521} {Generating senses
  and roles: An end-to-end model for dependency- and span-based semantic role
  labeling}.
\newblock In \emph{Proceedings of IJCAI}, pages 3786--3793.

\bibitem[{Cai et~al.(2018)Cai, He, Li, and Zhao}]{cai-etal-2018-full}
Jiaxun Cai, Shexia He, Zuchao Li, and Hai Zhao. 2018.
\newblock \href {https://aclanthology.org/C18-1233} {A full end-to-end semantic
  role labeler, syntactic-agnostic over syntactic-aware?}
\newblock In \emph{Proceedings of COLING}, pages 2753--2765, Santa Fe, New
  Mexico, USA.

\bibitem[{Cai et~al.(2017)Cai, Jiang, and Tu}]{cai-etal-2017-crf}
Jiong Cai, Yong Jiang, and Kewei Tu. 2017.
\newblock \href {https://doi.org/10.18653/v1/D17-1171} {{CRF} autoencoder for
  unsupervised dependency parsing}.
\newblock In \emph{Proceedings of EMNLP}, pages 1638--1643, Copenhagen,
  Denmark.

\bibitem[{Carreras and M{\`a}rquez(2005)}]{carreras-marquez-2005-introduction}
Xavier Carreras and Llu{\'\i}s M{\`a}rquez. 2005.
\newblock \href {https://aclanthology.org/W05-0620} {Introduction to the
  {C}o{NLL}-2005 shared task: Semantic role labeling}.
\newblock In \emph{Proceedings of CoNLL}, pages 152--164, Ann Arbor, Michigan.

\bibitem[{Choi and Palmer(2010)}]{choi-palmer-2010-retrieving}
Jinho Choi and Martha Palmer. 2010.
\newblock \href {https://aclanthology.org/W10-1811} {Retrieving correct
  semantic boundaries in dependency structure}.
\newblock In \emph{Proceedings of LAW}, pages 91--99, Uppsala, Sweden.

\bibitem[{Christensen et~al.(2010)Christensen, {Mausam}, Soderland, and
  Etzioni}]{christensen-etal-2010-semantic}
Janara Christensen, {Mausam}, Stephen Soderland, and Oren Etzioni. 2010.
\newblock \href {https://aclanthology.org/W10-0907} {Semantic role labeling for
  open information extraction}.
\newblock In \emph{Proceedings of WS}, pages 52--60, Los Angeles, California.

\bibitem[{Chu et~al.(2017)Chu, Jiang, and Tu}]{chu-etal-2017-latent}
Shanbo Chu, Yong Jiang, and Kewei Tu. 2017.
\newblock \href {https://ojs.aaai.org/index.php/AAAI/article/view/11047}
  {Latent dependency forest models}.
\newblock In \emph{Proceedings of AAAI}, pages 3733--3739.

\bibitem[{Cohn and Blunsom(2005)}]{cohn-blunsom-2005-semantic}
Trevor Cohn and Philip Blunsom. 2005.
\newblock \href {https://aclanthology.org/W05-0622} {Semantic role labelling
  with tree conditional random fields}.
\newblock In \emph{Proceedings of {C}o{NLL}}, pages 169--172, Ann Arbor,
  Michigan.

\bibitem[{Collins(2003)}]{collins-2003-head}
Michael Collins. 2003.
\newblock \href {https://doi.org/10.1162/089120103322753356} {Head-driven
  statistical models for natural language parsing}.
\newblock \emph{CL}, pages 589--637.

\bibitem[{Conia and Navigli(2020)}]{conia-navigli-2020-bridging}
Simone Conia and Roberto Navigli. 2020.
\newblock \href {https://doi.org/10.18653/v1/2020.coling-main.120} {Bridging
  the gap in multilingual semantic role labeling: a language-agnostic
  approach}.
\newblock In \emph{Proceedings of COLING}, pages 1396--1410, Barcelona, Spain
  (Online).

\bibitem[{Devlin et~al.(2019)Devlin, Chang, Lee, and
  Toutanova}]{devlin-etal-2019-bert}
Jacob Devlin, Ming-Wei Chang, Kenton Lee, and Kristina Toutanova. 2019.
\newblock \href {https://doi.org/10.18653/v1/N19-1423} {{BERT}: Pre-training of
  deep bidirectional transformers for language understanding}.
\newblock In \emph{Proceedings of NAACL-HLT}, pages 4171--4186, Minneapolis,
  Minnesota.

\bibitem[{Dozat and Manning(2017)}]{dozat-etal-2017-biaffine}
Timothy Dozat and Christopher~D. Manning. 2017.
\newblock \href {https://openreview.net/forum?id=Hk95PK9le} {Deep biaffine
  attention for neural dependency parsing}.
\newblock In \emph{Proceedings of ICLR}, Toulon, France.

\bibitem[{Eisner(1996)}]{eisner-1996-three}
Jason Eisner. 1996.
\newblock \href {https://aclanthology.org/C96-1058} {Three new probabilistic
  models for dependency parsing: An exploration}.
\newblock In \emph{Proceedings of COLING}, pages 340--345.

\bibitem[{Eisner(2000)}]{eisner-2000-bilexical}
Jason Eisner. 2000.
\newblock \href {https://www.cs.jhu.edu/~jason/papers/eisner.iwptbook00.pdf}
  {\emph{Bilexical Grammars and their Cubic-Time Parsing Algorithms}}, pages
  29--61. Dordrecht.

\bibitem[{Eisner(2016)}]{eisner-2016-inside}
Jason Eisner. 2016.
\newblock \href {https://doi.org/10.18653/v1/W16-5901} {Inside-outside and
  forward-backward algorithms are just backprop (tutorial paper)}.
\newblock In \emph{Proceedings of WS}, pages 1--17, Austin, TX.

\bibitem[{Eisner and Satta(1999)}]{eisner-satta-1999-efficient}
Jason Eisner and Giorgio Satta. 1999.
\newblock \href {https://doi.org/10.3115/1034678.1034748} {Efficient parsing
  for bilexical context-free grammars and head automaton grammars}.
\newblock In \emph{Proceedings of ACL}, pages 457--464, College Park, Maryland,
  USA.

\bibitem[{FitzGerald et~al.(2015)FitzGerald, T{\"a}ckstr{\"o}m, Ganchev, and
  Das}]{fitzgerald-etal-2015-semantic}
Nicholas FitzGerald, Oscar T{\"a}ckstr{\"o}m, Kuzman Ganchev, and Dipanjan Das.
  2015.
\newblock \href {https://doi.org/10.18653/v1/D15-1112} {Semantic role labeling
  with neural network factors}.
\newblock In \emph{Proceedings of EMNLP}, pages 960--970, Lisbon, Portugal.

\bibitem[{Fonseca and Martins(2020)}]{fonseca-martins-2020-revisiting}
Erick Fonseca and Andr{\'e} F.~T. Martins. 2020.
\newblock \href {https://doi.org/10.18653/v1/2020.acl-main.776} {Revisiting
  higher-order dependency parsers}.
\newblock In \emph{Proceedings of ACL}, pages 8795--8800, Online.

\bibitem[{Fu et~al.(2021)Fu, Tan, Chen, Huang, and Huang}]{fu-etal-2021-nested}
Yao Fu, Chuanqi Tan, Mosha Chen, Songfang Huang, and Fei Huang. 2021.
\newblock \href {https://ojs.aaai.org/index.php/AAAI/article/view/17519}
  {Nested named entity recognition with partially-observed treecrfs}.
\newblock In \emph{Proceedings of AAAI}, pages 12839--12847, Online.

\bibitem[{Gal and Ghahramani(2016)}]{yarin-etal-2016-dropout}
Yarin Gal and Zoubin Ghahramani. 2016.
\newblock \href {http://proceedings.mlr.press/v48/gal16.html} {Dropout as a
  bayesian approximation: Representing model uncertainty in deep learning}.
\newblock In \emph{Proceedings of ICML}, pages 1050--1059.

\bibitem[{Gildea and Hockenmaier(2003)}]{gildea-hockenmaier-2003-identifying}
Daniel Gildea and Julia Hockenmaier. 2003.
\newblock \href {https://aclanthology.org/W03-1008} {Identifying semantic roles
  using {C}ombinatory {C}ategorial {G}rammar}.
\newblock In \emph{Proceedings of EMNLP}, pages 57--64.

\bibitem[{Gildea and Jurafsky(2000)}]{gildea-jurafsky-2002-automatic}
Daniel Gildea and Daniel Jurafsky. 2000.
\newblock \href {https://doi.org/10.3115/1075218.1075283} {Automatic labeling
  of semantic roles}.
\newblock In \emph{Proceedings of ACL}, pages 512--520, Hong Kong.

\bibitem[{Gormley et~al.(2014)Gormley, Mitchell, Van~Durme, and
  Dredze}]{gormley-etal-2014-low}
Matthew~R. Gormley, Margaret Mitchell, Benjamin Van~Durme, and Mark Dredze.
  2014.
\newblock \href {https://doi.org/10.3115/v1/P14-1111} {Low-resource semantic
  role labeling}.
\newblock In \emph{Proceedings of ACL}, pages 1177--1187, Baltimore, Maryland.

\bibitem[{Hacioglu(2004)}]{hacioglu-2004-semantic}
Kadri Hacioglu. 2004.
\newblock \href {https://aclanthology.org/C04-1186} {Semantic role labeling
  using dependency trees}.
\newblock In \emph{Proceedings of COLING}, pages 1273--1276, Geneva,
  Switzerland.

\bibitem[{Haji{\v{c}} et~al.(2009)Haji{\v{c}}, Ciaramita, Johansson, Kawahara,
  Mart{\'\i}, M{\`a}rquez, Meyers, Nivre, Pad{\'o}, {\v{S}}t{\v{e}}p{\'a}nek,
  Stra{\v{n}}{\'a}k, Surdeanu, Xue, and Zhang}]{hajic-etal-2009-conll}
Jan Haji{\v{c}}, Massimiliano Ciaramita, Richard Johansson, Daisuke Kawahara,
  Maria~Ant{\`o}nia Mart{\'\i}, Llu{\'\i}s M{\`a}rquez, Adam Meyers, Joakim
  Nivre, Sebastian Pad{\'o}, Jan {\v{S}}t{\v{e}}p{\'a}nek, Pavel
  Stra{\v{n}}{\'a}k, Mihai Surdeanu, Nianwen Xue, and Yi~Zhang. 2009.
\newblock \href {https://aclanthology.org/W09-1201} {The {C}o{NLL}-2009 shared
  task: Syntactic and semantic dependencies in multiple languages}.
\newblock In \emph{Proceedings of {C}o{NLL}}, pages 1--18, Boulder, Colorado.

\bibitem[{Han et~al.(2017)Han, Jiang, and Tu}]{han-etal-2017-dependency}
Wenjuan Han, Yong Jiang, and Kewei Tu. 2017.
\newblock \href {https://doi.org/10.18653/v1/D17-1176} {Dependency grammar
  induction with neural lexicalization and big training data}.
\newblock In \emph{Proceedings of EMNLP}, pages 1683--1688, Copenhagen,
  Denmark.

\bibitem[{He and Choi(2021)}]{he-choi-2021-stem}
Han He and Jinho~D. Choi. 2021.
\newblock \href {https://aclanthology.org/2021.emnlp-main.451} {The stem cell
  hypothesis: Dilemma behind multi-task learning with transformer encoders}.
\newblock In \emph{Proceedings of EMNLP}, pages 5555--5577, Online and Punta
  Cana, Dominican Republic.

\bibitem[{He et~al.(2018{\natexlab{a}})He, Lee, Levy, and
  Zettlemoyer}]{he-etal-2018-jointly}
Luheng He, Kenton Lee, Omer Levy, and Luke Zettlemoyer. 2018{\natexlab{a}}.
\newblock \href {https://doi.org/10.18653/v1/P18-2058} {Jointly predicting
  predicates and arguments in neural semantic role labeling}.
\newblock In \emph{Proceedings of ACL}, pages 364--369, Melbourne, Australia.

\bibitem[{He et~al.(2017)He, Lee, Lewis, and Zettlemoyer}]{he-etal-2017-deep}
Luheng He, Kenton Lee, Mike Lewis, and Luke Zettlemoyer. 2017.
\newblock \href {https://doi.org/10.18653/v1/P17-1044} {Deep semantic role
  labeling: What works and what{'}s next}.
\newblock In \emph{Proceedings of ACL}, pages 473--483, Vancouver, Canada.

\bibitem[{He et~al.(2018{\natexlab{b}})He, Li, Zhao, and
  Bai}]{he-etal-2018-syntax}
Shexia He, Zuchao Li, Hai Zhao, and Hongxiao Bai. 2018{\natexlab{b}}.
\newblock \href {https://doi.org/10.18653/v1/P18-1192} {Syntax for semantic
  role labeling, to be, or not to be}.
\newblock In \emph{Proceedings of ACL}, pages 2061--2071, Melbourne, Australia.

\bibitem[{Henderson et~al.(2008)Henderson, Merlo, Musillo, and
  Titov}]{henderson-etal-2008-latent}
James Henderson, Paola Merlo, Gabriele Musillo, and Ivan Titov. 2008.
\newblock \href {https://aclanthology.org/W08-2122} {A latent variable model of
  synchronous parsing for syntactic and semantic dependencies}.
\newblock In \emph{Proceedings of CoNLL}, pages 178--182, Manchester, England.

\bibitem[{Henderson et~al.(2013)Henderson, Merlo, Titov, and
  Musillo}]{henderson-etal-2013-multilingual}
James Henderson, Paola Merlo, Ivan Titov, and Gabriele Musillo. 2013.
\newblock \href {https://doi.org/10.1162/COLI_a_00158} {Multilingual joint
  parsing of syntactic and semantic dependencies with a latent variable model}.
\newblock \emph{CL}, pages 949--998.

\bibitem[{Jiang et~al.(2019)Jiang, Li, Zhang, and Zhang}]{jiang-etal-2019-hlt}
Wei Jiang, Zhenghua Li, Yu~Zhang, and Min Zhang. 2019.
\newblock \href {https://doi.org/10.18653/v1/S19-2002} {{HLT}@{SUDA} at
  {S}em{E}val-2019 task 1: {UCCA} graph parsing as constituent tree parsing}.
\newblock In \emph{Proceedings of SemEval}, pages 11--15, Minneapolis,
  Minnesota, USA.

\bibitem[{Jiang et~al.(2016)Jiang, Han, and Tu}]{jiang-etal-2016-unsupervised}
Yong Jiang, Wenjuan Han, and Kewei Tu. 2016.
\newblock \href {https://doi.org/10.18653/v1/D16-1073} {Unsupervised neural
  dependency parsing}.
\newblock In \emph{Proceedings of EMNLP}, pages 763--771, Austin, Texas.

\bibitem[{Jie and Lu(2019)}]{jie-lu-2019-dependency}
Zhanming Jie and Wei Lu. 2019.
\newblock \href {https://doi.org/10.18653/v1/D19-1399} {Dependency-guided
  {LSTM}-{CRF} for named entity recognition}.
\newblock In \emph{Proceedings of EMNLP-IJCNLP}, pages 3862--3872, Hong Kong,
  China.

\bibitem[{Jindal et~al.(2020)Jindal, Aharonov, Brahma, Zhu, and
  Li}]{jindal-etal-2020-improved}
Ishan Jindal, Ranit Aharonov, Siddhartha Brahma, Huaiyu Zhu, and Yunyao Li.
  2020.
\newblock \href {http://arxiv.org/abs/2011.14459} {Improved semantic role
  labeling using parameterized neighborhood memory adaptation}.

\bibitem[{Johansson and
  Nugues(2008{\natexlab{a}})}]{johansson-nugues-2008-dependency}
Richard Johansson and Pierre Nugues. 2008{\natexlab{a}}.
\newblock \href {https://aclanthology.org/D08-1008} {Dependency-based semantic
  role labeling of {P}rop{B}ank}.
\newblock In \emph{Proceedings of EMNLP}, pages 69--78, Honolulu, Hawaii.

\bibitem[{Johansson and
  Nugues(2008{\natexlab{b}})}]{johansson-nugues-2008-dependency-based}
Richard Johansson and Pierre Nugues. 2008{\natexlab{b}}.
\newblock \href {https://aclanthology.org/W08-2123} {Dependency-based
  syntactic{--}semantic analysis with {P}rop{B}ank and {N}om{B}ank}.
\newblock In \emph{Proceedings of CoNLL}, pages 183--187, Manchester, England.

\bibitem[{Johansson and
  Nugues(2008{\natexlab{c}})}]{johansson-nugues-2008-effect}
Richard Johansson and Pierre Nugues. 2008{\natexlab{c}}.
\newblock \href {https://aclanthology.org/C08-1050} {The effect of syntactic
  representation on semantic role labeling}.
\newblock In \emph{Proceedings of COLING}, pages 393--400, Manchester, UK.

\bibitem[{Kim et~al.(2017)Kim, Denton, Hoang, and Rush}]{kim-2017-structured}
Yoon Kim, Carl Denton, Luong Hoang, and Alexander~M. Rush. 2017.
\newblock \href {https://openreview.net/forum?id=HkE0Nvqlg} {Structured
  attention networks}.
\newblock In \emph{Proceedings of ICLR}.

\bibitem[{Kingma and Ba(2015)}]{kingma-ba-2015-adam}
Diederik~P. Kingma and Jimmy Ba. 2015.
\newblock \href {http://arxiv.org/abs/1412.6980} {Adam: {A} method for
  stochastic optimization}.
\newblock In \emph{Proceedings of ICLR}, San Diego, CA, USA.

\bibitem[{Klein and Manning(2004)}]{klein-manning-2004-corpus}
Dan Klein and Christopher Manning. 2004.
\newblock \href {https://doi.org/10.3115/1218955.1219016} {Corpus-based
  induction of syntactic structure: Models of dependency and constituency}.
\newblock In \emph{Proceedings of ACL}, pages 478--485, Barcelona, Spain.

\bibitem[{Kong et~al.(2015)Kong, Rush, and Smith}]{kong-etal-2015-transforming}
Lingpeng Kong, Alexander~M. Rush, and Noah~A. Smith. 2015.
\newblock \href {https://doi.org/10.3115/v1/N15-1080} {Transforming
  dependencies into phrase structures}.
\newblock In \emph{Proceedings of NAACL-HLT}, pages 788--798, Denver, Colorado.

\bibitem[{Lafferty et~al.(2001)Lafferty, McCallum, and
  Pereira}]{lafferty-etal-2001-crf}
John~D. Lafferty, Andrew McCallum, and Fernando C.~N. Pereira. 2001.
\newblock \href {http://www.aladdin.cs.cmu.edu/papers/pdfs/y2001/crf.pdf}
  {Conditional random fields: Probabilistic models for segmenting and labeling
  sequence data}.
\newblock In \emph{Proceedings of ICML}, pages 282--289, Williams College,
  Williamstown, MA, USA.

\bibitem[{Lample et~al.(2016)Lample, Ballesteros, Subramanian, Kawakami, and
  Dyer}]{lample-etal-2016-neural}
Guillaume Lample, Miguel Ballesteros, Sandeep Subramanian, Kazuya Kawakami, and
  Chris Dyer. 2016.
\newblock \href {https://doi.org/10.18653/v1/N16-1030} {Neural architectures
  for named entity recognition}.
\newblock In \emph{Proceedings of NAACL-HLT}, pages 260--270, San Diego,
  California.

\bibitem[{Le and Zuidema(2015)}]{le-zuidema-2015-unsupervised}
Phong Le and Willem Zuidema. 2015.
\newblock \href {https://doi.org/10.3115/v1/N15-1067} {Unsupervised dependency
  parsing: Let{'}s use supervised parsers}.
\newblock In \emph{Proceedings of NAACL-HLT}, pages 651--661, Denver, Colorado.

\bibitem[{Li et~al.(2020)Li, Jawale, Palmer, and
  Srikumar}]{li-etal-2020-structured}
Tao Li, Parth~Anand Jawale, Martha Palmer, and Vivek Srikumar. 2020.
\newblock \href {https://doi.org/10.18653/v1/2020.acl-main.744} {Structured
  tuning for semantic role labeling}.
\newblock In \emph{Proceedings of ACL}, pages 8402--8412, Online.

\bibitem[{Li et~al.(2016)Li, Zhang, Zhang, Liu, Chen, Wu, and
  Wang}]{li-etal-2016-active}
Zhenghua Li, Min Zhang, Yue Zhang, Zhanyi Liu, Wenliang Chen, Hua Wu, and
  Haifeng Wang. 2016.
\newblock \href {https://doi.org/10.18653/v1/P16-1033} {Active learning for
  dependency parsing with partial annotation}.
\newblock In \emph{Proceedings of ACL}, pages 344--354, Berlin, Germany.

\bibitem[{Li et~al.(2019)Li, He, Zhao, Zhang, Zhang, Zhou, and
  Zhou}]{li-etal-2019-dependency}
Zuchao Li, Shexia He, Hai Zhao, Yiqing Zhang, Zhuosheng Zhang, Xi~Zhou, and
  Xiang Zhou. 2019.
\newblock \href {https://doi.org/10.1609/aaai.v33i01.33016730} {Dependency or
  span, end-to-end uniform semantic role labeling}.
\newblock In \emph{Proceedings of AAAI}, pages 6730--6737.

\bibitem[{Li et~al.(2021)Li, Zhao, He, and Cai}]{li-etal-2021-syntax}
Zuchao Li, Hai Zhao, Shexia He, and Jiaxun Cai. 2021.
\newblock \href
  {https://direct.mit.edu/coli/article/47/3/529/102778/Syntax-Role-for-Neural-Semantic-Role-Labeling}
  {Syntax role for neural semantic role labeling}.
\newblock \emph{CL}, pages 529--574.

\bibitem[{Lin et~al.(2017)Lin, Liu, and Sun}]{lin-etal-2017-neural}
Yankai Lin, Zhiyuan Liu, and Maosong Sun. 2017.
\newblock \href {https://doi.org/10.18653/v1/P17-1004} {Neural relation
  extraction with multi-lingual attention}.
\newblock In \emph{Proceedings of ACL}, pages 34--43, Vancouver, Canada.

\bibitem[{Liu and Gildea(2010)}]{liu-gildea-2010-semantic}
Ding Liu and Daniel Gildea. 2010.
\newblock \href {https://aclanthology.org/C10-1081} {Semantic role features for
  machine translation}.
\newblock In \emph{Proceedings of COLING}, pages 716--724, Beijing, China.

\bibitem[{Liu et~al.(2022)Liu, Jiang, Cotterell, and
  Sachan}]{liu-etal-2022-structured}
Tianyu Liu, Yuchen Jiang, Ryan Cotterell, and Mrinmaya Sachan. 2022.
\newblock \href {https://aclanthology.org/2022.naacl-main.189} {A structured
  span selector}.
\newblock In \emph{Proceedings of NAACL}, pages 2629--2641, Seattle, United
  States.

\bibitem[{Liu and Lapata(2018)}]{liu-lapata-2018-learning}
Yang Liu and Mirella Lapata. 2018.
\newblock \href {https://doi.org/10.1162/tacl_a_00005} {Learning structured
  text representations}.
\newblock \emph{TACL}, pages 63--75.

\bibitem[{Loshchilov and Hutter(2019)}]{ilya-etal-2018-adamw}
Ilya Loshchilov and Frank Hutter. 2019.
\newblock \href {https://openreview.net/forum?id=Bkg6RiCqY7} {Decoupled weight
  decay regularization}.
\newblock In \emph{Proceedings of ICLR}, New Orleans, LA, USA.

\bibitem[{Lou et~al.(2022)Lou, Yang, and Tu}]{lou-etal-2022-nested}
Chao Lou, Songlin Yang, and Kewei Tu. 2022.
\newblock \href {https://aclanthology.org/2022.acl-long.428} {Nested named
  entity recognition as latent lexicalized constituency parsing}.
\newblock In \emph{Proceedings of ACL}, pages 6183--6198, Dublin, Ireland.

\bibitem[{Marcheggiani and Titov(2017)}]{marcheggiani-titov-2017-encoding}
Diego Marcheggiani and Ivan Titov. 2017.
\newblock \href {https://doi.org/10.18653/v1/D17-1159} {Encoding sentences with
  graph convolutional networks for semantic role labeling}.
\newblock In \emph{Proceedings of EMNLP}, pages 1506--1515, Copenhagen,
  Denmark.

\bibitem[{Marcus et~al.(1993)Marcus, Santorini, and
  Marcinkiewicz}]{marcus-etal-1993-building}
Mitchell~P. Marcus, Beatrice Santorini, and Mary~Ann Marcinkiewicz. 1993.
\newblock \href {https://aclanthology.org/J93-2004} {Building a large annotated
  corpus of {E}nglish: The {P}enn {T}reebank}.
\newblock \emph{CL}, pages 313--330.

\bibitem[{McDonald et~al.(2005)McDonald, Crammer, and
  Pereira}]{mcdonald-etal-2005-online}
Ryan McDonald, Koby Crammer, and Fernando Pereira. 2005.
\newblock \href {https://doi.org/10.3115/1219840.1219852} {Online large-margin
  training of dependency parsers}.
\newblock In \emph{Proceedings of ACL}, pages 91--98, Ann Arbor, Michigan.

\bibitem[{McDonald and Pereira(2006)}]{mcdonald-pereira-2006-online}
Ryan McDonald and Fernando Pereira. 2006.
\newblock \href {https://aclanthology.org/E06-1011} {Online learning of
  approximate dependency parsing algorithms}.
\newblock In \emph{Proceedings of EACL}, pages 81--88, Trento, Italy.

\bibitem[{Meila and Jordan(2000)}]{marina-michael-2000-mixure}
Marina Meila and Michael~I. Jordan. 2000.
\newblock \href {http://jmlr.org/papers/v1/meila00a.html} {Learning with
  mixtures of trees}.
\newblock \emph{JMLR}, pages 1--48.

\bibitem[{Mohammadshahi and Henderson(2021)}]{mohammadshahi-etal-2021-g2g}
Alireza Mohammadshahi and James Henderson. 2021.
\newblock \href {http://arxiv.org/abs/2104.07704} {Syntax-aware graph-to-graph
  transformer for semantic role labelling}.

\bibitem[{Naradowsky et~al.(2012)Naradowsky, Riedel, and
  Smith}]{naradowsky-etal-2012-improving}
Jason Naradowsky, Sebastian Riedel, and David Smith. 2012.
\newblock \href {https://aclanthology.org/D12-1074} {Improving {NLP} through
  marginalization of hidden syntactic structure}.
\newblock In \emph{Proceedings of EMNLP}, pages 810--820, Jeju Island, Korea.

\bibitem[{Nivre et~al.(2014)Nivre, Goldberg, and
  McDonald}]{nivre-etal-2014-squibs}
Joakim Nivre, Yoav Goldberg, and Ryan McDonald. 2014.
\newblock \href {https://doi.org/10.1162/COLI_a_00184} {{S}quibs: Constrained
  arc-eager dependency parsing}.
\newblock \emph{CL}, pages 249--257.

\bibitem[{Ouchi et~al.(2018)Ouchi, Shindo, and
  Matsumoto}]{ouchi-etal-2018-span}
Hiroki Ouchi, Hiroyuki Shindo, and Yuji Matsumoto. 2018.
\newblock \href {https://doi.org/10.18653/v1/D18-1191} {A span selection model
  for semantic role labeling}.
\newblock In \emph{Proceedings of EMNLP}, pages 1630--1642, Brussels, Belgium.

\bibitem[{Palmer et~al.(2005)Palmer, Gildea, and
  Kingsbury}]{palmer-etal-2005-propbank}
Martha Palmer, Daniel Gildea, and Paul Kingsbury. 2005.
\newblock \href {https://doi.org/10.1162/0891201053630264} {The {P}roposition
  {B}ank: An annotated corpus of semantic roles}.
\newblock \emph{CL}, pages 71--106.

\bibitem[{Paolini et~al.(2021)Paolini, Athiwaratkun, Krone, Ma, Achille,
  ANUBHAI, dos Santos, Xiang, and Soatto}]{paolini-etal-2021-structured}
Giovanni Paolini, Ben Athiwaratkun, Jason Krone, Jie Ma, Alessandro Achille,
  RISHITA ANUBHAI, Cicero~Nogueira dos Santos, Bing Xiang, and Stefano Soatto.
  2021.
\newblock \href {https://openreview.net/forum?id=US-TP-xnXI} {Structured
  prediction as translation between augmented natural languages}.
\newblock In \emph{Proceedings of ICLR}.

\bibitem[{Pereira and Warren(1983)}]{pereira-warren-1983-parsing}
Fernando C.~N. Pereira and David H.~D. Warren. 1983.
\newblock \href {https://doi.org/10.3115/981311.981338} {Parsing as deduction}.
\newblock In \emph{Proceedings of ACL}, pages 137--144, Cambridge,
  Massachusetts, USA.

\bibitem[{Pradhan et~al.(2013)Pradhan, Moschitti, Xue, Ng, Bj{\"o}rkelund,
  Uryupina, Zhang, and Zhong}]{pradhan-etal-2013-towards}
Sameer Pradhan, Alessandro Moschitti, Nianwen Xue, Hwee~Tou Ng, Anders
  Bj{\"o}rkelund, Olga Uryupina, Yuchen Zhang, and Zhi Zhong. 2013.
\newblock \href {https://aclanthology.org/W13-3516} {Towards robust linguistic
  analysis using {O}nto{N}otes}.
\newblock In \emph{Proceedings of CoNLL-WS}, pages 143--152, Sofia, Bulgaria.

\bibitem[{Pradhan et~al.(2012)Pradhan, Moschitti, Xue, Uryupina, and
  Zhang}]{pradhan-etal-2012-conll}
Sameer Pradhan, Alessandro Moschitti, Nianwen Xue, Olga Uryupina, and Yuchen
  Zhang. 2012.
\newblock \href {https://aclanthology.org/W12-4501} {{C}o{NLL}-2012 shared
  task: Modeling multilingual unrestricted coreference in {O}nto{N}otes}.
\newblock In \emph{Proceedings of CoNLL-WS}, pages 1--40, Jeju Island, Korea.

\bibitem[{Punyakanok et~al.(2008)Punyakanok, Roth, and
  Yih}]{punyakanok-etal-2008-importance}
Vasin Punyakanok, Dan Roth, and Wen-tau Yih. 2008.
\newblock \href {https://doi.org/10.1162/coli.2008.34.2.257} {The importance of
  syntactic parsing and inference in semantic role labeling}.
\newblock \emph{CL}, pages 257--287.

\bibitem[{Punyakanok et~al.(2004)Punyakanok, Roth, Yih, and
  Zimak}]{punyakanok-etal-2004-semantic}
Vasin Punyakanok, Dan Roth, Wen-tau Yih, and Dav Zimak. 2004.
\newblock \href {https://aclanthology.org/C04-1197} {Semantic role labeling via
  integer linear programming inference}.
\newblock In \emph{Proceedings of COLING}, pages 1346--1352, Geneva,
  Switzerland.

\bibitem[{Rush(2020)}]{rush-2020-torch}
Alexander Rush. 2020.
\newblock \href {https://doi.org/10.18653/v1/2020.acl-demos.38} {Torch-struct:
  Deep structured prediction library}.
\newblock In \emph{Proceedings of ACL}, pages 335--342, Online.

\bibitem[{Shen et~al.(2021)Shen, Tay, Zheng, Bahri, Metzler, and
  Courville}]{shen-etal-2021-structformer}
Yikang Shen, Yi~Tay, Che Zheng, Dara Bahri, Donald Metzler, and Aaron
  Courville. 2021.
\newblock \href {https://doi.org/10.18653/v1/2021.acl-long.559}
  {{S}truct{F}ormer: Joint unsupervised induction of dependency and
  constituency structure from masked language modeling}.
\newblock In \emph{Proceedings of ACL-IJCNLP}, pages 7196--7209, Online.

\bibitem[{Shi and Lin(2019)}]{shi-etal-2019-simple}
Peng Shi and Jimmy Lin. 2019.
\newblock \href {http://arxiv.org/abs/1904.05255} {Simple bert models for
  relation extraction and semantic role labeling}.

\bibitem[{Shi et~al.(2020)Shi, Malioutov, and Irsoy}]{shi-etal-2020-semantic}
Tianze Shi, Igor Malioutov, and Ozan Irsoy. 2020.
\newblock \href {https://doi.org/10.18653/v1/2020.emnlp-main.610} {Semantic
  role labeling as syntactic dependency parsing}.
\newblock In \emph{Proceedings of EMNLP}, pages 7551--7571, Online.

\bibitem[{Smith and Eisner(2008)}]{smith-eisner-2008-dependency}
David Smith and Jason Eisner. 2008.
\newblock \href {https://aclanthology.org/D08-1016} {Dependency parsing by
  belief propagation}.
\newblock In \emph{Proceedings of EMNLP}, pages 145--156, Honolulu, Hawaii.

\bibitem[{Strubell et~al.(2018)Strubell, Verga, Andor, Weiss, and
  McCallum}]{strubell-etal-2018-lisa}
Emma Strubell, Patrick Verga, Daniel Andor, David Weiss, and Andrew McCallum.
  2018.
\newblock \href {https://doi.org/10.18653/v1/D18-1548} {Linguistically-informed
  self-attention for semantic role labeling}.
\newblock In \emph{Proceedings of EMNLP}, pages 5027--5038, Brussels, Belgium.

\bibitem[{Sun et~al.(2017)Sun, Cao, and Wan}]{sun-etal-2017-semantic}
Weiwei Sun, Junjie Cao, and Xiaojun Wan. 2017.
\newblock \href {https://doi.org/10.18653/v1/P17-1077} {Semantic dependency
  parsing via book embedding}.
\newblock In \emph{Proceedings of ACL}, pages 828--838, Vancouver, Canada.

\bibitem[{Swayamdipta et~al.(2018)Swayamdipta, Thomson, Lee, Zettlemoyer, Dyer,
  and Smith}]{swayamdipta-etal-2018-syntactic}
Swabha Swayamdipta, Sam Thomson, Kenton Lee, Luke Zettlemoyer, Chris Dyer, and
  Noah~A. Smith. 2018.
\newblock \href {https://doi.org/10.18653/v1/D18-1412} {Syntactic scaffolds for
  semantic structures}.
\newblock In \emph{Proceedings of EMNLP}, pages 3772--3782, Brussels, Belgium.

\bibitem[{T{\"a}ckstr{\"o}m et~al.(2015)T{\"a}ckstr{\"o}m, Ganchev, and
  Das}]{tackstrom-etal-2015-efficient}
Oscar T{\"a}ckstr{\"o}m, Kuzman Ganchev, and Dipanjan Das. 2015.
\newblock \href {https://doi.org/10.1162/tacl_a_00120} {Efficient inference and
  structured learning for semantic role labeling}.
\newblock \emph{TACL}, pages 29--41.

\bibitem[{Tan et~al.(2018)Tan, Wang, Xie, Chen, and Shi}]{tan-etal-2018-deep}
Zhixing Tan, Mingxuan Wang, Jun Xie, Yidong Chen, and Xiaodong Shi. 2018.
\newblock \href
  {https://www.aaai.org/ocs/index.php/AAAI/AAAI18/paper/view/16725} {Deep
  semantic role labeling with self-attention}.
\newblock In \emph{Proceedings of AAAI}, pages 4929--4936.

\bibitem[{Toutanova et~al.(2008)Toutanova, Haghighi, and
  Manning}]{toutanova-etal-2008-global}
Kristina Toutanova, Aria Haghighi, and Christopher~D. Manning. 2008.
\newblock \href {https://doi.org/10.1162/coli.2008.34.2.161} {A global joint
  model for semantic role labeling}.
\newblock \emph{CL}, pages 161--191.

\bibitem[{Vaswani et~al.(2017)Vaswani, Shazeer, Parmar, Uszkoreit, Jones,
  Gomez, Kaiser, and Polosukhin}]{vaswani-2017-attention}
Ashish Vaswani, Noam Shazeer, Niki Parmar, Jakob Uszkoreit, Llion Jones,
  Aidan~N. Gomez, Lukasz Kaiser, and Illia Polosukhin. 2017.
\newblock \href
  {https://proceedings.neurips.cc/paper/2017/hash/3f5ee243547dee91fbd053c1c4a845aa-Abstract.html}
  {Attention is all you need}.
\newblock In \emph{Advances in NIPS}, pages 5998--6008, Long Beach, California,
  USA.

\bibitem[{Wang et~al.(2019)Wang, Huang, and Tu}]{wang-etal-2019-second}
Xinyu Wang, Jingxian Huang, and Kewei Tu. 2019.
\newblock \href {https://doi.org/10.18653/v1/P19-1454} {Second-order semantic
  dependency parsing with end-to-end neural networks}.
\newblock In \emph{Proceedings of ACL}, pages 4609--4618, Florence, Italy.

\bibitem[{Xia et~al.(2019)Xia, Li, Zhang, Zhang, Fu, Wang, and
  Si}]{xia-etal-2019-syntax}
Qingrong Xia, Zhenghua Li, Min Zhang, Meishan Zhang, Guohong Fu, Rui Wang, and
  Luo Si. 2019.
\newblock \href {https://ojs.aaai.org/index.php/AAAI/article/view/4717}
  {Syntax-aware neural semantic role labeling}.
\newblock In \emph{Proceedings of AAAI}, pages 7305--7313.

\bibitem[{Xu et~al.(2021)Xu, Jie, Lu, and Bing}]{xu-etal-2021-better}
Lu~Xu, Zhanming Jie, Wei Lu, and Lidong Bing. 2021.
\newblock \href {https://doi.org/10.18653/v1/2021.naacl-main.271} {Better
  feature integration for named entity recognition}.
\newblock In \emph{Proceedings of NAACL-HLT}, pages 3457--3469, Online.

\bibitem[{Yang et~al.(2020)Yang, Jiang, Han, and Tu}]{yang-etal-2020-second}
Songlin Yang, Yong Jiang, Wenjuan Han, and Kewei Tu. 2020.
\newblock \href {https://doi.org/10.18653/v1/2020.coling-main.347}
  {Second-order unsupervised neural dependency parsing}.
\newblock In \emph{Proceedings of COLING}, pages 3911--3924, Barcelona, Spain
  (Online).

\bibitem[{Yang and Tu(2022)}]{yang-tu-2022-combining}
Songlin Yang and Kewei Tu. 2022.
\newblock \href {https://aclanthology.org/2022.findings-acl.112} {Combining
  (second-order) graph-based and headed-span-based projective dependency
  parsing}.
\newblock In \emph{Findings of ACL}, pages 1428--1434, Dublin, Ireland.

\bibitem[{Yang et~al.(2021)Yang, Zhao, and Tu}]{yang-etal-2021-neural}
Songlin Yang, Yanpeng Zhao, and Kewei Tu. 2021.
\newblock \href {https://doi.org/10.18653/v1/2021.acl-long.209} {Neural
  bi-lexicalized {PCFG} induction}.
\newblock In \emph{Proceedings of ACL-IJCNLP}, pages 2688--2699, Online.

\bibitem[{Yih et~al.(2016)Yih, Richardson, Meek, Chang, and
  Suh}]{yih-etal-2016-value}
Wen-tau Yih, Matthew Richardson, Chris Meek, Ming-Wei Chang, and Jina Suh.
  2016.
\newblock \href {https://doi.org/10.18653/v1/P16-2033} {The value of semantic
  parse labeling for knowledge base question answering}.
\newblock In \emph{Proceedings of ACL}, pages 201--206, Berlin, Germany.

\bibitem[{Yu et~al.(2020)Yu, Bohnet, and Poesio}]{yu-etal-2020-named}
Juntao Yu, Bernd Bohnet, and Massimo Poesio. 2020.
\newblock \href {https://doi.org/10.18653/v1/2020.acl-main.577} {Named entity
  recognition as dependency parsing}.
\newblock In \emph{Proceedings of ACL}, pages 6470--6476, Online.

\bibitem[{Zhang et~al.(2021{\natexlab{a}})Zhang, Wang, Han, and
  Tu}]{zhang-etal-2021-adapting}
Liwen Zhang, Ge~Wang, Wenjuan Han, and Kewei Tu. 2021{\natexlab{a}}.
\newblock \href {https://doi.org/10.18653/v1/2021.acl-long.449} {Adapting
  unsupervised syntactic parsing methodology for discourse dependency parsing}.
\newblock In \emph{Proceedings of ACL-IJCNLP}, pages 5782--5794, Online.

\bibitem[{Zhang et~al.(2019{\natexlab{a}})Zhang, Ma, Duh, and
  Van~Durme}]{zhang-etal-2019-amr}
Sheng Zhang, Xutai Ma, Kevin Duh, and Benjamin Van~Durme. 2019{\natexlab{a}}.
\newblock \href {https://doi.org/10.18653/v1/P19-1009} {{AMR} parsing as
  sequence-to-graph transduction}.
\newblock In \emph{Proceedings of ACL}, pages 80--94, Florence, Italy.

\bibitem[{Zhang et~al.(2019{\natexlab{b}})Zhang, Ma, Duh, and
  Van~Durme}]{zhang-etal-2019-broad}
Sheng Zhang, Xutai Ma, Kevin Duh, and Benjamin Van~Durme. 2019{\natexlab{b}}.
\newblock \href {https://doi.org/10.18653/v1/D19-1392} {Broad-coverage semantic
  parsing as transduction}.
\newblock In \emph{Proceedings of EMNLP-IJCNLP}, pages 3786--3798, Hong Kong,
  China.

\bibitem[{Zhang et~al.(2020)Zhang, Li, and Zhang}]{zhang-etal-2020-efficient}
Yu~Zhang, Zhenghua Li, and Min Zhang. 2020.
\newblock \href {https://doi.org/10.18653/v1/2020.acl-main.302} {Efficient
  second-order {T}ree{CRF} for neural dependency parsing}.
\newblock In \emph{Proceedings of ACL}, pages 3295--3305, Online.

\bibitem[{Zhang et~al.(2021{\natexlab{b}})Zhang, Strubell, and
  Hovy}]{zhang-etal-2021-comparing}
Zhisong Zhang, Emma Strubell, and Eduard Hovy. 2021{\natexlab{b}}.
\newblock \href {https://doi.org/10.18653/v1/2021.spnlp-1.8} {Comparing span
  extraction methods for semantic role labeling}.
\newblock In \emph{Proceedings of SPNLP}, pages 67--77, Online.

\bibitem[{Zhou and Xu(2015)}]{zhou-xu-2015-end}
Jie Zhou and Wei Xu. 2015.
\newblock \href {https://doi.org/10.3115/v1/P15-1109} {End-to-end learning of
  semantic role labeling using recurrent neural networks}.
\newblock In \emph{Proceedings of ACL-IJCNLP}, pages 1127--1137, Beijing,
  China.

\bibitem[{Zhou et~al.(2020{\natexlab{a}})Zhou, Li, and
  Zhao}]{zhou-etal-2020-parsing}
Junru Zhou, Zuchao Li, and Hai Zhao. 2020{\natexlab{a}}.
\newblock \href {https://doi.org/10.18653/v1/2020.findings-emnlp.398} {Parsing
  all: Syntax and semantics, dependencies and spans}.
\newblock In \emph{Findings of EMNLP}, pages 4438--4449, Online.

\bibitem[{Zhou et~al.(2020{\natexlab{b}})Zhou, Zhang, Zhao, and
  Zhang}]{zhou-etal-2020-limit}
Junru Zhou, Zhuosheng Zhang, Hai Zhao, and Shuailiang Zhang.
  2020{\natexlab{b}}.
\newblock \href {https://doi.org/10.18653/v1/2020.findings-emnlp.399}
  {{LIMIT}-{BERT} : Linguistics informed multi-task {BERT}}.
\newblock In \emph{Findings of EMNLP}, pages 4450--4461, Online.

\bibitem[{Zhou et~al.(2020{\natexlab{c}})Zhou, Zhang, Ji, and
  Tang}]{zhou-etal-2020-amr}
Qiji Zhou, Yue Zhang, Donghong Ji, and Hao Tang. 2020{\natexlab{c}}.
\newblock \href {https://doi.org/10.18653/v1/2020.acl-main.397} {{AMR} parsing
  with latent structural information}.
\newblock In \emph{Proceedings of ACL}, pages 4306--4319, Online.

\bibitem[{Zhou et~al.(2022)Zhou, Xia, Li, Zhang, and
  Zhang}]{zhou-etal-2022-fast}
Shilin Zhou, Qingrong Xia, Zhenghua Li, Yu~Zhang, and Min Zhang. 2022.
\newblock \href {https://arxiv.org/abs/2112.02970} {Fast and accurate
  span-based semantic role labeling as graph parsing}.
\newblock In \emph{Proceedings of COLING}.

\bibitem[{Zhou et~al.(2020{\natexlab{d}})Zhou, Jiang, Hu, and
  Tu}]{zhou-etal-2020-latent}
Yang Zhou, Yong Jiang, Zechuan Hu, and Kewei Tu. 2020{\natexlab{d}}.
\newblock \href {http://arxiv.org/abs/2011.05009} {Neural latent dependency
  model for sequence labeling}.

\bibitem[{Zhu et~al.(2020)Zhu, Bisk, and Neubig}]{zhu-etal-2020-return}
Hao Zhu, Yonatan Bisk, and Graham Neubig. 2020.
\newblock \href {https://doi.org/10.1162/tacl_a_00337} {The return of lexical
  dependencies: Neural lexicalized {PCFG}s}.
\newblock \emph{TACL}, pages 647--661.

\end{thebibliography}

\appendix

\section{Implementation Details}\label{sec:impl}

In this work, we set up two alternative model architectures, i.e, LSTM-based and PLM-based.
For the LSTM-based model, we directly adopt most settings of \citet{dozat-etal-2017-biaffine} with some adaptions.
The input vector of each token $x_i\in\boldsymbol{x}$ is the concatenation of three parts,
\begin{equation*}
    \mathbf{e}_i = \left[\mathbf{e}^{\mathrm{word}}_i;\mathbf{e}^{\mathrm{lemma}}_i;\mathbf{e}^{\mathrm{char}}_i\right]
\end{equation*}
where $\mathbf{e}_i^{\mathrm{word}}$ and $\mathbf{e}_i^{\mathrm{lemma}}$ are word and lemma embeddings, and $\mathbf{e}^{\mathrm{char}}_i$ is the outputs of a CharLSTM layer \cite{lample-etal-2016-neural}.
We set the dimension of lemma and CharLSTM representations to 100 in our setting.
We next feed the input embeddings into 3-layer BiLSTMs \cite{yarin-etal-2016-dropout} to get contextualized representations with dimension 800.
\begin{equation*}
    \mathbf{h}_0,\mathbf{h}_1,\dots,\mathbf{h}_n=\mathtt{BiLSTMs}(\mathbf{e}_0,\mathbf{e}_1,\dots,\mathbf{e}_n)
\end{equation*}
Other dimension settings are kept the same as biaffine parser \cite{dozat-etal-2017-biaffine}.
Following \citet{zhang-etal-2020-efficient}, we set the hidden size of Triaffine layer to 100 for \textsc{Crf2o} additionally.
The training process continues at most 1,000 epochs and is early stopped if the performance on Dev data does not increase in 100 consecutive epochs.
In practice, we observe that the training procedure is often stopped within 300 epochs ($\sim$12 hours), which is efficient enough.

For PLM-based models, we opt to directly finetune the PLM layers without cascading word embedding and LSTM layers for the sake of simplicity.
We use ``\href{https://huggingface.co/bert-large-cased}{\texttt{bert-large-cased}}'' for BERT, and ``\href{https://huggingface.co/roberta-large}{\texttt{roberta-large}}'' for RoBERTa respectively.
We train the model for 20 epochs with roughly 1,000 tokens per batch and use AdamW \cite{kingma-ba-2015-adam,ilya-etal-2018-adamw} with $\beta_1=0.9,\beta_2=0.9$ and $\lambda=0$ for parameter optimization .
The learning rate is $5\times 10^{-5}$ for PLMs, and $10^{-3}$ for the rest components.
We adopt the warmup strategy in the first 10\% of the training steps, and then apply a linear decay to the learning rate in the remaining steps.

\section{The Inside Algorithm}\label{sec:inside}

We give the pseudocode of the common second-order Inside algorithm \cite{mcdonald-pereira-2006-online} in Alg.~\ref{alg:inside-2o} as additional explanations to Fig.~\ref{fig:deduction}.
The difference between the common second-order Inside algorithm and our proposed span-constrained one lies in the rule constraints green highlighted in Fig.~\ref{fig:deduction}.
\begin{algorithm}[tb!]
    \begin{algorithmic}[1]
        \newlength{\commentindent}
        \setlength{\commentindent}{.24\textwidth}
        \renewcommand{\algorithmiccomment}[1]{\unskip\hfill\makebox[\commentindent][l]{$\rhd$~#1}\par}
        \LetLtxMacro{\oldalgorithmic}{\algorithmic}
        \renewcommand{\algorithmic}[1][0]{%
            \oldalgorithmic[#1]
            \renewcommand{\ALC@com}[1]{\ifnum\pdfstrcmp{##1}{default}=0\else\algorithmiccomment{##1}\fi}
        }
        \STATE \textbf{Define:} $I, S, C \in \mathbf{R}^{n \times n \times b}$
        \STATE $\;$ \COMMENT{$b$ is batch size}
        \STATE \textbf{Initialize:} $C_{i, i} = \mathbf{0}, 0 \le i \le n$\label{alg:init}
        \FOR [span width]{$w = 1$ \TO $n$}
        \STATE \emph{Parallelization on} $0 \le i$; $j=i+w \le n$
        \STATE $\begin{aligned}
                I_{i, j} & \leftarrow \log(\exp(C_{i, i} + C_{j, i+1})                         \\
                         & + \sum_{i < r < j} \exp(I_{i, r} + S_{r, j} + \mathrm{s}(i, r, j))) \\
                         & + \mathrm{s}(i, j)
            \end{aligned} $\label{alg:incomplete-r}
        \STATE $\begin{aligned}
                I_{j, i} & \leftarrow \log(\exp(C_{j, j} + C_{i, j-1})                         \\
                         & + \sum_{i < r < j} \exp(I_{j, r} + S_{r, i} + \mathrm{s}(i, r, j))) \\
                         & + \mathrm{s}(j, i)
            \end{aligned} $\label{alg:incomplete-l}

        \STATE $S_{i, j} \leftarrow \log\sum_{i \le r < j} \exp(C_{i, r} +  C_{j, r+1})$\label{alg:sib}
        \STATE $C_{i, j} \leftarrow \log\sum_{i < r \le j} \exp(I_{i, r} +  C_{r, j})$\label{alg:complete-r}
        \STATE $C_{j, i} \leftarrow \log\sum_{i \le r < j} \exp(I_{j, r} +  C_{r, i})$\label{alg:complete-l}
        \ENDFOR
        \RETURN $C_{0, n}$
    \end{algorithmic}
    \caption{The Second-order Inside Algorithm.}
    \label{alg:inside-2o}
\end{algorithm}

In Line~\ref{alg:init}, $C_{i,i}$ corresponds to the axiom items $\tikz[baseline=-10pt]{\righttriangle[0.6][0.3]{i}{i}{}}$ with initial score $\mathbf{0}$.
Line~\ref{alg:incomplete-r} corresponds to two merge operations in Fig.~\ref{fig:deduction}.
The incomplete span $I_{i,j}$ ($\tikz[baseline=-10pt]{\trapezoid[0.6][0.3][0.1]{i}{j}{}}$) is obtained by summing over either all pairs of complete span $C_{i,i}$ and $C_{j,i+1}$ (\textbf{\textsc{R-Link}}) or pairs of the incomplete span $I_{i,r}$ and the sibling span $S_{j,r}$ (\textbf{\textsc{R-Link2}}).
In Line~\ref{alg:sib}, the sibling span $S_{i,j}$ ($\tikz[baseline=-10pt]{\square[0.6][0.3]{i}{j}{}}$) is obtained by summing over all pairs of complete span $C_{i,r}$ and $C_{j,r+1}$ (\textbf{\textsc{Comb}}).
Line~\ref{alg:complete-r} describes the similar merging operation on all pairs of the incomplete span $I_{i,r}$ and the complete span $C_{r,j}$, resulting a complete span $C_{i,j}$ ($\tikz[baseline=-10pt]{\righttriangle[0.6][0.3]{i}{j}{}}$) (\textbf{\textsc{R-Comb}}).
Line~\ref{alg:incomplete-l} and Line~\ref{alg:complete-l} is the symmetric L-rules, which are omitted in Fig.~\ref{fig:deduction}.

\section{More Comparisons}\label{sec:incomparable}

\begin{table}[tb!]
    \renewcommand{\arraystretch}{1.1}
    \setlength{\tabcolsep}{2.7pt}
    \centering
    \begin{small}
        \begin{tabular}{l ccc ccc}
            \toprule
            \rowcolor[gray]{0.95}                    & \multicolumn{3}{c}{WSJ} & \multicolumn{3}{c}{Brown}                                                                 \\
            \rulefiller	\cmidrule(lr){2-4}		\cmidrule(lr){5-7}
            \rowcolor[gray]{0.95}                    & P                       & R                         & F$_1$         & P             & R             & F$_1$         \\
            \midrule
            SA$^\diamondsuit$                        & 84.17                   & 83.28                     & 83.72         & 72.98         & 70.1\white{0} & 71.51         \\
            SA$^{\diamondsuit}_\texttt{ELMo}$        & 86.21                   & 85.98                     & 86.09         & 77.1\white{0} & 75.61         & 76.35         \\
            G2G$^\clubsuit_\texttt{BERT}$            & 86.40                   & 87.79                     & 87.08         & 78.76         & 80.06         & 79.40         \\
            LIMIT$^\clubsuit_\texttt{BERT}$          & 86.62                   & 89.12                     & 87.85         & 79.58         & 83.05         & 81.28         \\
            ParsingAll$^\clubsuit_\texttt{BERT}$     & 86.77                   & 88.49                     & 87.62         & 79.06         & 81.67         & 80.34         \\
            ParsingAll$^\clubsuit_\texttt{XLNet}$    & 87.65                   & 89.66                     & 88.64         & 80.77         & 83.92         & 82.31         \\\\[-10pt]
            \textsc{Crf}\rlap{$_\texttt{BERT}$}      & 86.98                   & 88.28                     & 87.63         & 79.19         & 80.92         & 80.05         \\
            \textsc{Crf2o}\rlap{$_\texttt{BERT}$}    & 87.00                   & 88.76                     & 87.87         & 79.08         & 81.50         & 80.27         \\
            \textsc{Crf}\rlap{$_\texttt{RoBERTa}$}   & 87.20                   & 88.67                     & 87.93         & 79.29         & 81.48         & 80.38         \\
            \textsc{Crf2o}\rlap{$_\texttt{RoBERTa}$} & 87.35                   & 89.34                     & 88.33         & 79.95         & 82.32         & 81.12         \\
            \rowcolor[gray]{0.95}\multicolumn{7}{c}{\emph{w/ gold predicates}}                                                                                             \\
            ParsingAll$^\clubsuit_\texttt{BERT}$     & 89.04                   & 88.79                     & 88.91         & 81.89         & 80.98         & 81.43         \\
            ParsingAll$^\clubsuit_\texttt{XLNet}$    & 89.89                   & 89.74                     & 89.81         & 85.35         & 84.57         & 84.96         \\
            TANL$^\diamondsuit_\texttt{T5}$          & -                       & -                         & 89.3\white{0} & -             & -             & 82.0\white{0} \\\\[-10pt]
            \textsc{Crf}\rlap{$_\texttt{BERT}$}      & 88.93                   & 88.58                     & 88.76         & 82.87         & 81.67         & 82.27         \\
            \textsc{Crf2o}\rlap{$_\texttt{BERT}$}    & 89.00                   & 89.03                     & 89.02         & 82.81         & 82.35         & 82.58         \\
            \textsc{Crf}\rlap{$_\texttt{RoBERTa}$}   & 89.29                   & 88.99                     & 89.15         & 83.22         & 82.42         & 82.82         \\
            \textsc{Crf2o}\rlap{$_\texttt{RoBERTa}$} & 89.45                   & 89.63                     & 89.54         & 83.89         & 83.39         & 83.64         \\
            \bottomrule
        \end{tabular}
        \caption{
            Comparisons with other less comparable works on CoNLL05 WSJ and Brown data.
            $^\clubsuit$ means using linguistic syntax knowledge;
            $^\diamondsuit$ means different evaluation methods.
            SA: \citet{strubell-etal-2018-lisa}; ParsingAll: \citet{zhou-etal-2020-parsing}; LIMIT: \citet{zhou-etal-2020-limit}; G2G: \citet{mohammadshahi-etal-2021-g2g}; TANL: \citet{paolini-etal-2021-structured}.
        }
        \label{table:incomparable}
    \end{small}
\end{table}

In Table~\ref{table:incomparable}, for reference, we list the results of some works with different experimental settings and therefore less comparable.
For example, \citet{paolini-etal-2021-structured} and \citet{strubell-etal-2018-lisa}\footnote{
    Under the \emph{end-to-end} setting, different from the standard pratice \cite{he-etal-2018-jointly}, \citet{strubell-etal-2018-lisa} only ran the evaluation tool once, resulting in slightly higher precision values.
    See discussions in \href{https://github.com/strubell/LISA/issues/9}{their code issue}.
}
adopt different evaluation metrics, resulting in slightly higher F$_1$ values than official tools.
Nonetheless, we find that our \textsc{Crf2o} with RoBERTa achieves 89.54 F$_1$ on WSJ data under the \emph{w/ gold predicates} setting, showing very competitive results when compared with T5-based model of \citet{paolini-etal-2021-structured}.
\citet{zhou-etal-2020-parsing} propose a joint-learning framework, integrating both (dependency/constituency) syntactic parse trees and dependency-based SRL resources to enhance their models.
Their ablation studies show that using syntax trees brought an overall improvement of 1.6 F$_1$ score on CoNLL05 Dev data.
We believe that we could achieve similar or even higher results than their syntax-aware XLNet-based models by incorporating human-annotated syntax knowledge.
However, exploring different ways of injecting syntax is not the core of this paper.
We take this as our future work.

\begin{table}[tb!]
    \renewcommand{\arraystretch}{1.1}
    \centering
    \small
    \begin{tabular}{l ccc}
        \toprule
        \rowcolor[gray]{0.95}                           & P              & R              & F$_1$          \\
        \midrule
        \textsc{Crf}                                    & 75.28          & 75.24          & 75.26          \\
        \textsc{Crf}$_\texttt{BERT}$                    & 84.70          & 84.39          & 84.54          \\
        \rowcolor[gray]{0.95} \multicolumn{4}{c}{\emph{w/ gold syntax}}                                    \\
        \citet{johansson-nugues-2008-effect}            & -              & -              & 84.32          \\
        \citet{li-etal-2019-dependency}$_\texttt{ELMo}$ & -              & -              & 89.20          \\
        \textsc{Crf}$_\texttt{BERT}$                    & \textbf{93.56} & \textbf{93.22} & \textbf{93.39} \\
        \bottomrule
    \end{tabular}
    \caption{Results for dependency-based evaluation on CoNLL09 Test data under \emph{w/o.} and \emph{w/ gold syntax} settings.}
    \label{table:conll09}
\end{table}

\section{Dependency-based evaluation}\label{sec:tree-probing}

Observing that our \textsc{Crf} model can conveniently determine dependencies from predicates to span headwords as by-products of constructing arguments, we therefore conduct dependency-based evaluation on CoNLL09 Test data \cite{hajic-etal-2009-conll} to measure the quality of induced dependencies.
As CoNLL09 data shares the same text content with CoNLL05, we directly make use of the model trained on CoNLL05 to obtain the results of CoNLL09 Test.
Following \citet{johansson-nugues-2008-dependency-based,li-etal-2019-dependency}, we also compare our \textsc{Crf} outputs with the upper bound of utilizing gold syntax tree to determine the headwords of predicted arguments.
Since CoNLL05 contains only verbal predicates, we discard all nominal predicate-argument structures under the guidance of POS tags starting with \texttt{N*}.
Word senses and self-loops are removed as well.

Results are listed in Table~\ref{table:conll09}, from which we can draw some observations:
1) after using BERT, \textsc{Crf} outperforms LSTM-based model (75.26) by a large margin, implying BERT provides fruitful prior knowledge for dependency induction;
2) our \textsc{Crf} with BERT achieves 84.54 F$_1$ on CoNLL09 Test, exhibiting very promising performance even when compared to models using gold syntax \cite{johansson-nugues-2008-dependency-based,li-etal-2019-dependency}.
This indicates that the dependencies induced by \textsc{Crf} are highly in line with gold dependency-based annotations, illuminating potential extensions of our work on supervised dependency-based SRL.

\section{Grammar Induction}\label{sec:induction}

\begin{table}[tb!]
    \renewcommand{\arraystretch}{1.1}
    \setlength{\tabcolsep}{5pt}
    \centering
    \begin{small}
        \begin{tabular}{lll}
            \toprule
            \rowcolor[gray]{0.95} Rules & Models                                          & WSJ           \\
            \midrule
            \multirow{5}{*}{Stanford}   & NL-PCFGs \cite{zhu-etal-2020-return}            & 40.5          \\
                                        & NBL-PCFGs \cite{yang-etal-2021-neural}          & 39.1          \\
                                        & StructFormer \cite{shen-etal-2021-structformer} & 46.2          \\\\[-10pt]
                                        & \textsc{Crf}                                    & 48.0          \\
                                        & \textsc{Crf}$_\texttt{BERT}$                    & \textbf{65.4} \\
            \rowcolor[gray]{0.95}\multicolumn{3}{c}{\emph{w/ gold POS tags (for reference)}}              \\
            \multirow{6}{*}{Collins}    & DMV \cite{klein-manning-2004-corpus}            & 39.4          \\
                                        & MaxEnc \cite{le-zuidema-2015-unsupervised}      & 65.8          \\
                                        & NDMV \cite{jiang-etal-2016-unsupervised}        & 57.6          \\
                                        & CRFAE \cite{cai-etal-2017-crf}                  & 55.7          \\
                                        & L-NDMV \cite{han-etal-2017-dependency}          & 59.5          \\
                                        & NDMV2o \cite{yang-etal-2020-second}             & \textbf{67.5} \\
            \bottomrule
        \end{tabular}
        \caption{Grammar induction results of our \textsc{Crf} model under different head-finding rules.}
        \label{table:induction}
    \end{small}
\end{table}

To gain further insights, we make use of the scores defined in Eq.~\ref{eq:tree-score} to extract full dependency tree structures.
Surprisingly, we find they are highly in agreement with expert-designed grammars \cite{marcus-etal-1993-building} when examined on the grammar induction task \cite{klein-manning-2004-corpus}.

We show precise grammar induction results in Table~\ref{table:induction}.
The results are not comparable to typical methods like DMV \cite{klein-manning-2004-corpus} or CRFAE \cite{cai-etal-2017-crf}, as they use gold POS tags as guidance, and we use Stanford Dependencies rather than Collins rules \cite{collins-2003-head}.
Under similar settings, however, our learned task-specific trees perform significantly better than recent works.

Another interesting observation is that the gap between the BERT-based model and the LSTM-based model is much larger than that on SRL results.
This implies LSTMs tend to be more fitted to SRL structures, while BERT is able to provide a strong inductive bias for syntax induction.

\end{document}